\documentclass[10pt,twocolumn,letterpaper]{article}

\usepackage[pagenumbers]{iccv} % To force page numbers, e.g. for an arXiv version

\usepackage{bm}
\definecolor{iccvblue}{rgb}{0.21,0.49,0.74}
\usepackage[pagebackref,breaklinks,colorlinks,allcolors=iccvblue]{hyperref}

\usepackage[accsupp]{axessibility}  % Improves PDF readability for those with disabilities.

\usepackage{transparent}
\usepackage{pifont}
\usepackage{bbding}
\usepackage{color}
\usepackage{xcolor}
\usepackage{colortbl}
\usepackage{wrapfig}
\usepackage{graphicx}

\usepackage{fmtcount}

\usepackage{fontawesome5}
\usepackage[misc]{ifsym}

\usepackage{soul}  % background text color

\definecolor{lightblue}{RGB}{224,239,251}
\definecolor{lightpurple}{RGB}{227,227,241}
\definecolor{gray60}{gray}{.60}
\definecolor{gray92}{gray}{.92}
\definecolor{gray94}{gray}{.94}
\definecolor{gray96}{gray}{.96}
\newcommand{\grow}[1]{\rowcolor{gray94}{#1}} % row as gray

\newcommand{\brow}[1]{\rowcolor{lightblue}{#1}} % row as blue

\newcommand{\xmarkg}{\textcolor{gray60}{\XSolidBrush}}

\definecolor{resultblue}{rgb}{0.21,0.49,0.76}

\definecolor{repmtlred}{RGB}{206, 100, 111}
\colorlet{repmtlred}{repmtlred!111}

\definecolor{repmtlblue}{RGB}{96, 132, 192}
\colorlet{repmtlblue}{repmtlblue!111}

\def\paperID{4340} % *** Enter the Paper ID here
\def\confName{ICCV}
\def\confYear{2025}

\title{Rep-MTL: Unleashing the Power of Representation-level Task Saliency \\ for Multi-Task Learning}

\author{
\begin{tabular}[t]{@{}c@{}}
Zedong Wang$^1$ \quad Siyuan Li$^2$ \quad Dan Xu$^{1, }${\textsuperscript{\Letter}}
\end{tabular}\\[1ex]
\begin{tabular}[t]{@{}c@{}}
$^1$The Hong Kong University of Science and Technology \quad $^2$Zhejiang University
\end{tabular}\\[0.5ex]
{\tt\small zedong.wang@connect.ust.hk, lisiyuan@westlake.edu.cn, danxu@cse.ust.hk
}}

\begin{document}
\maketitle
\begin{abstract}
Despite the promise of Multi-Task Learning in leveraging complementary knowledge across tasks, existing multi-task optimization (MTO) techniques remain fixated on resolving conflicts via optimizer-centric loss scaling and gradient manipulation strategies, yet fail to deliver consistent gains. In this paper, we argue that the shared representation space, where task interactions naturally occur, offers rich information and potential for operations complementary to existing optimizers, especially for facilitating the inter-task complementarity, which is rarely explored in MTO. This intuition leads to Rep-MTL, which exploits the representation-level task saliency to quantify interactions between task-specific optimization and shared representation learning. By steering these saliencies through entropy-based penalization and sample-wise cross-task alignment, Rep-MTL aims to mitigate negative transfer by maintaining the effective training of individual tasks instead pure conflict-solving, while explicitly promoting complementary information sharing. Experiments are conducted on four challenging MTL benchmarks covering both task-shift and domain-shift scenarios. The results show that Rep-MTL, even paired with the basic equal weighting policy, achieves competitive performance gains with favorable efficiency. Beyond standard performance metrics, Power Law exponent analysis demonstrates Rep-MTL's efficacy in balancing task-specific learning and cross-task sharing. The project page is available at \href{https://jacky1128.github.io/RepMTL/}{HERE}.

\end{abstract}    
\section{Introduction}
\label{sec:intro}

Multi-Task Learning (MTL)~\citep{caruana1997MultiTask} has garnered increasing attention in recent years, with notable success in computer vision~\citep{doersch2017MTLVision, kirillov2023SAM}, natural language processing~\citep{radford2019MTLNLP, bubeck2023sparks}, and other modalities~\citep{lu2022UnifiedIO, lu2024UnifiedIO2}. By leveraging multiple supervision signals at once, MTL models are expected to learn robust representation with reduced cost but better generalization compared to their single-task learning (STL) counterparts~\citep{icml24PlatonicRep}.

\vspace{0.05em}
However, performance deterioration occurs as the tasks involved may not necessarily exhibit significant correlation, which induces conflicts in joint training~\citep{yu2020PCGrad, chen2020GradDrop}, \ie, competing updates to the same architecture could impede the effective training of individual tasks, thus resulting in inferior convergence and worse generalization~\citep{kendall2018UW, zhang2022NegativeTransfer}. This hence makes optimization an integral part of MTL, with its crux believed to be alleviating the \textit{negative transfer} among tasks while exploiting their \textit{positive complementarity}~\citep{standley2020ICML, xin2022current, shen2024GO4Align}.

\begin{figure}[t]
    %\vspace{-0.5em}
    \centering
    \includegraphics[width=0.48\textwidth]{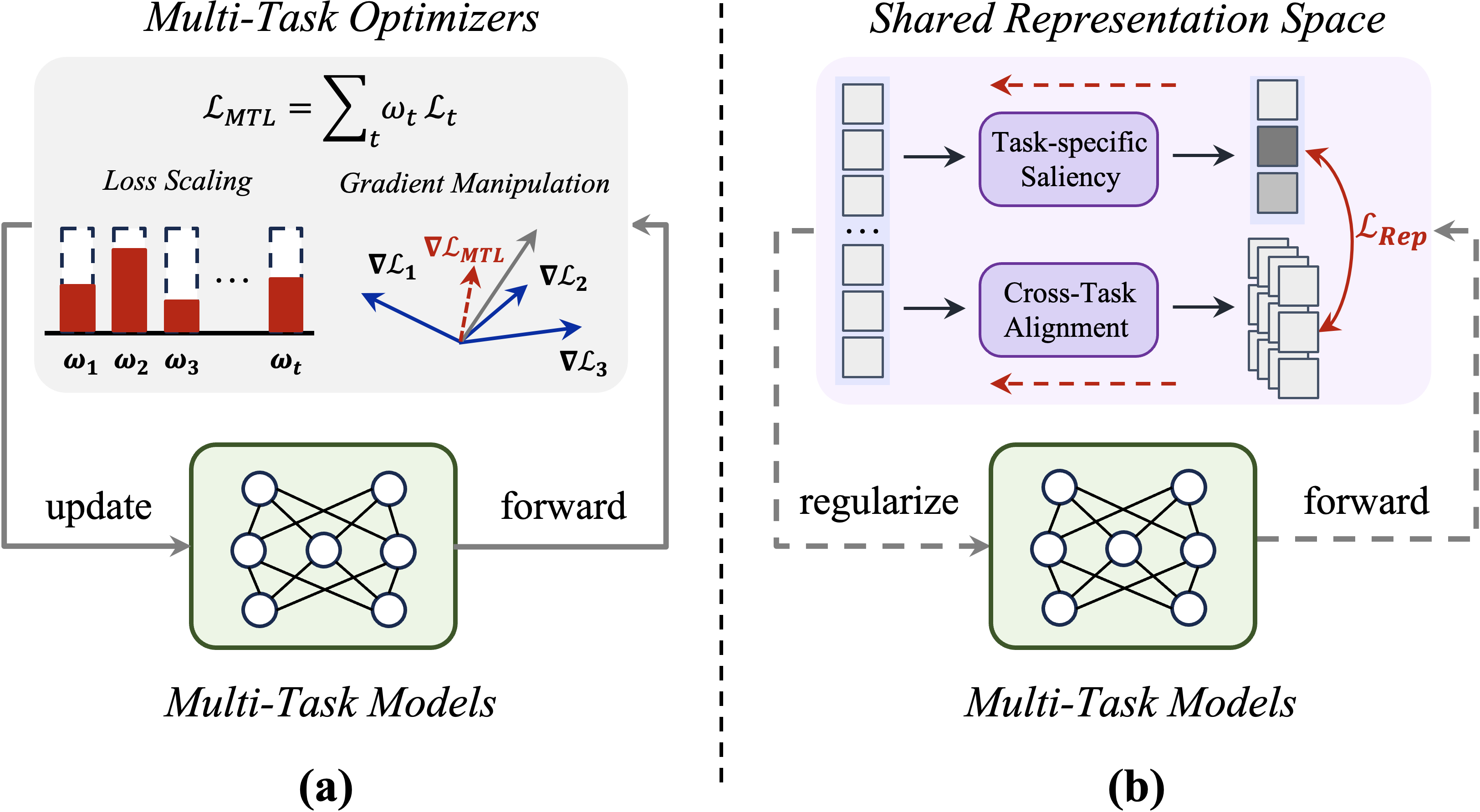}
    \vspace{-1.70em}
    \caption{Overview of Rep-MTL and existing MTO methods. (a) Both loss scaling and gradient manipulation focus on optimizer-centric policies to address conflicts in model update. (b) Rep-MTL exploits shared representation space to facilitate cross-task sharing while preserving task-specific signals without optimizer changes.
    }
    \label{fig:teaser}
    \vspace{-1.4em}
\end{figure}

\vspace{0.05em}
As such, various studies have focused on addressing negative transfer through optimizer-centric loss scaling~\citep{
liu2019DWA, liu2024FAMO} and gradient manipulation~\citep{navon2022NashMTL, icml24FairGrad}. Although more precise multi-task optimization (MTO) strategies have been introduced, their inconsistent efficacy has been increasingly recognized, especially in demanding scenarios~\citep{kurin2022defense, xin2022current, lin2023LibMTL, mueller2024MTLRep}. In parallel, the probe of task relationships has been extended into shared representation space~\citep{Javaloy2021RotoGradGH, KDD2022SRDML, mao2024FeatureDisentangle}, in which task saliencies are quantified by gradients \wrt shared representation~\citep{selvaraju2017GradCAM}, opening new avenues for MTO. In this paper, we argue that the representation space, where task interactions naturally occur, offers rich information and potential for operations beyond (and complementary to) optimizer designs, especially for facilitating task complementarity \textit{explicitly}.

\begin{figure*}[t!]
    \vspace{-0.5em}
    \centering
    \includegraphics[width=1.00\textwidth]{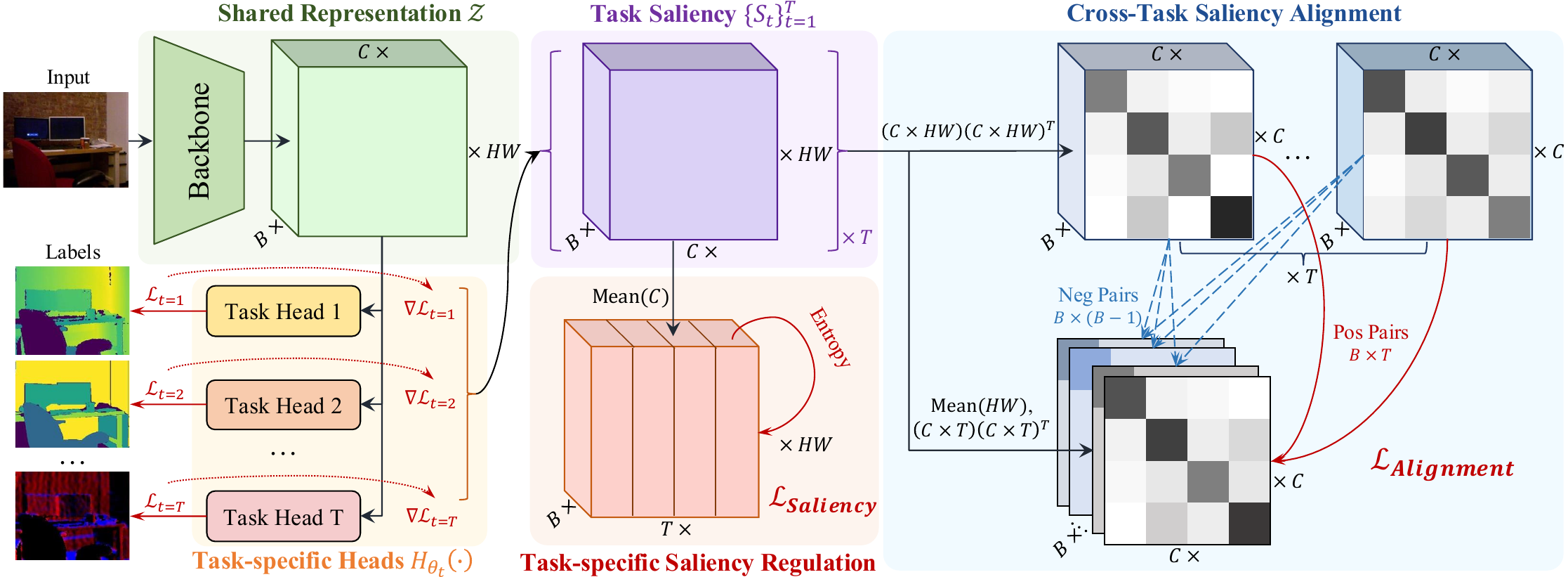}
    \vspace{-1.30em}
    \caption{The Rep-MTL framework. It comprises two complementary task saliency driven modules: \textit{(i) Task-specific Saliency Regulation (TSR)} that penalizes the saliency distribution to emphasize task-salient patterns during training, thereby mitigating negative transfer among tasks. \textit{(ii) Cross-task Saliency Alignment (CSA)} that facilitates inter-task complementarity by aligning saliencies while maintaining task distinctiveness through positive and negative pairs. Together, Rep-MTL enables transfer across tasks while keeping task-specific patterns.
    }
    \label{fig:framework}
    \vspace{-1.25em}
\end{figure*}

\vspace{0.05em}
To this end, we introduce Rep-MTL, a representation-centric approach that modulates multi-task training via task saliency (Sec.~\ref{subsec:formulation}). As shown in Figure~\ref{fig:framework}, it comprises two associative modules: (i) \underline{\textbf{T}}ask-specific \underline{\textbf{S}}aliency \underline{\textbf{R}}egulation (TSR) that penalizes the task saliency distributions through an entropy-based regularization, ensuring that task-specific patterns can be reserved and remain distinctive during training, thereby mitigating negative transfer in MTO. (ii) \underline{\textbf{C}}ross-task \underline{\textbf{S}}aliency \underline{\textbf{A}}lignment (CSA). Since prior methods rarely touched on this, we start with an intuitive question: \textit{What additional message does representation space provide us during training?} -- the rich sample-wise feature dimensions first come to our mind and inspire our design of CSA, which \textit{explicitly} promotes inter-task complementarity by aligning sample-wise saliencies in a contrastive learning paradigm. Together, Rep-MTL as a regularization requires no further modifications to either optimizers or network architectures.

As aforementioned, MTO methods might perform worse in demanding settings. To rigorously validate Rep-MTL's effectiveness, we evaluate it from three key aspects: First, we conduct experiments on four challenging MTL benchmarks that encompass both task-shift and domain-shift scenarios, where most MTO baselines exhibit negative performance gains. Second, we examine Rep-MTL's robustness and practical applicability by analyzing its sensitivity to hyper-parameters (Sec.~\ref{subsec:exp_hyper}), learning rates (Appendix~\ref{app:learning_rate}), and its optimization speed (Sec.~\ref{subsec:exp_runtime}). Third, beyond standard performance metrics, we apply Power Law (PL) exponent analysis~\citep{JMLR2021selfreg, NC2021WeightWatcher} to validate if and how Rep-MTL influences model updates (Sec.~\ref{subsection:exp_ple_analysis}). Specifically, it verifies how different trained model parts (\eg, backbone or decoders) fit their related objectives (overall or task-specific losses) under different MTO methods. 
Overall, the empirical results show that: (i) Rep-MTL consistently achieves competitive performance, even with the basic equal weighting (EW). (ii) Rep-MTL is more efficient than most gradient manipulation methods (\eg, $\sim$26\% and $\sim$12\% faster than Nash-MTL~\citep{navon2022NashMTL} and FairGrad~\citep{icml24FairGrad}, respectively). (iii) It successfully maintains effective training of individual tasks while exploiting inter-task complementarity for more robust MTL models.

Our contributions can thus be summarized as follows:
\begin{itemize}[leftmargin=1.15em]
\vspace{0.10em}
    \item 
    We introduce Rep-MTL, a representation-centric MTO approach that aims to mitigate negative transfer while exploiting inter-task complementarity with task saliency.
    \vspace{0.30em}
    \item 
    Rep-MTL as a regularization method complements existing MTO strategies, achieving competitive performance on diverse MTL benchmarks even with the basic EW.
    \vspace{0.30em}
    \item 
    Observation and insights are obtained from empirical evidence: (i) Mitigating negative transfer could go beyond pure conflict-solving strategies. TSR offers another path by ensuring the effective training of individual tasks. (ii) The explicit cross-task sharing in CSA has shown significant potential, but remains under-explored in MTO.
\end{itemize}

\section{Related Work}
\label{sec:related}

\subsection{Multi-Task Optimization}

Prior work in MTO operates under the assumption that the negative transfer stems from task conflicts in gradient directions and magnitude. Thus, three types of optimizer-centric methods have been developed: loss scaling, gradient manipulation, and hybrid ones. Below, we discuss where existing studies fall on each group and where our Rep-MTL stands.

\vspace{-1.00em}
\paragraph{Loss Scaling.}
Loss scaling methods optimize both shared and task-specific parameters by adjusting task-specific loss weights, evolving from EW~\citep{zhang2021survey} to more complex policies, such as homoscedastic uncertainty-based scaling~\citep{kendall2018UW}, and loss changing-based adaptation~\citep{liu2019DWA}.
GLS~\citep{chennupati2019GLS} minimizes the geometric mean loss, while IMTL-L~\citep{liu2021IMTL} proposes an adaptive loss transformation to maintain constant weighted losses. RLW~\citep{lin22RLW} employs probabilistic sampling from normal distributions, and IGBv2~\citep{dai2023IGBv2} utilizes improbable gaps for scale computation. More recently, FAMO~\citep{liu2024FAMO} balances different losses by decreasing them approximately at an equal rate. GO4Align~\citep{shen2024GO4Align} goes a step further through task grouping-based interaction for dynamic progress alignment.

\vspace{-1.00em}
\paragraph{Gradient Manipulation and Hybrid Methods.}
Gradient manipulation techniques adjust and aggregate task-specific gradients \wrt shared parameters to address conflicts. Early attempts focused on basic gradient modification: GradNorm~\citep{chen2018GradNorm} equalizes gradient magnitudes with learned task weights, whereas PCGrad~\citep{yu2020PCGrad} resolves conflicts by gradient projection. GradVac~\citep{wang2021GradVac} and GradDrop~\citep{chen2020GradDrop} introduce the universal alignment and consistency-based gradient sign dropout, respectively. MGDA~\citep{desideri12MGDA} for the first time views the MTO issue as a multi-objective optimization problem and pioneered the search for Pareto-optimal solutions. Subsequently, this paradigm was improved by CAGrad~\citep{liu2021CAGrad}, which optimizes for the worst-case task improvement, and MoCo~\citep{fernando2023MoCo}, which incorporates momentum estimation to eliminate bias. Nash-MTL~\citep{navon2022NashMTL} incorporates game theory to find Nash bargaining solutions, while IMTL-G~\citep{liu2021IMTL} updates gradient directions that all their cosine similarities are equal. Recon~\citep{Shi2023ReconRC} first employs Neural Architecture Search to deal with gradient conflicts. More recently, Aligned-MTL~\citep{senushkin2023AlignedMTL} leverages principal component alignment for stability. FairGrad~\citep{icml24FairGrad} and STCH~\citep{icml24STCH} reformulate MTO with utility maximization and smoothed Tchebycheff optimization.
Hybrid methods~\citep{liu2021IMTL, linreasonableTMLR2022, liuautoTMLR2022, lin2023DBMTL} combine the above two groups of technique to cash in their complementary strengths.

\vspace{-1.00em}
\paragraph{Shared Representation in MTO.} 
Several studies~\citep{sener2018multi, navon2022NashMTL, senushkin2023AlignedMTL} have utilized gradients \wrt shared representation rather than parameters to update models, which reduces the computational overhead with limited backpropagation through task-specific decoders. Among these, RotoGrad~\citep{Javaloy2021RotoGradGH} and SRDML~\citep{KDD2022SRDML} stand as the closest studies to Rep-MTL. RotoGrad employs feature rotation to minimize task semantic disparities, while SRDML learns task relations by regularizing task-wise similarity of saliencies. While they have made substantial progress, they solve task conflicts with task similarities and leave the essential inter-task complementarity entirely to the MTL architectures.

\subsection{MTL Architectures}

Hard parameter sharing (HPS) architecture~\citep{caruana1997MultiTask}, while fundamental for MTL, suffers from \textit{negative transfer} among tasks. Subsequent work has thus focused on facilitating cross-task positive transfer through various architecture designs: from learnable fusion weights (\eg, Cross-stitch~\citep{MisraSGH16CrossStitch}, Sluice network~\citep{ruder2019SluiceNet}), to dynamic expert combinations (\eg, MoE~\citep{shazeer2017MoE}, MMoE~\citep{ma2018MMoE}, PLE~\citep{tang2020PLE}), and attention-based feature extraction like MTAN~\citep{liu2019DWA}. Moreover, several studies explore knowledge transfer at output layer through sequential modeling (ESSM~\citep{ma2018ESSM}) or distillation (CrossDistil~\citep{yang2022CrossDistil}). While these architectures have proven effective in exploiting inter-task complementarity, MTO has mainly focused on addressing negative transfer by solving gradient conflicts. In this work, we aim to bridge this gap by introducing an MTO method that \textit{explicitly} facilitates such complementarity while maintaining the canonical HPS architecture.
\section{Methodology}

This section presents Rep-MTL, which advances multi-task learning through task saliency regularization in shared representation space. We first preview the MTO problem and representation-level task saliency (Sec.~\ref{subsec:formulation}). Subsequently, we propose two complementary components of Rep-MTL: (i) task-specific saliency regulation that aims to mitigate negative MTL transfer by preserving task-salient learning patterns (Sec.~\ref{subsec:saliency}), and (ii) sample-wise cross-task alignment for facilitating inter-task complementarity (Sec.~\ref{subsec:alignment}).

\subsection{Preliminaries and Task Saliency}
\label{subsec:formulation}
Consider a set of $T$ correlated tasks $\{\mathcal{T}_t\}_{t=1}^{T}$, where $T \in \mathbb{N}^+$ denotes the total number of tasks. In HPS architectures, model parameters $\theta = \{\theta_{s}, \{\theta_t\}_{t=1}^{T}\}$ comprise shared parameters $\theta_{s}$ for the backbone and task-specific parameters $\{\theta_t\}_{t=1}^{T}$ for decoders, where each $\theta_t$ corresponds to task $\mathcal{T}_t$. 

For illustration, we present the framework in the context of computer vision. Given an input RGB image $X \in \mathbb{R}^{3 \times H \times W}$, a shared encoder $E_{\theta_{s}}(\cdot)$ maps the input $X$ into a latent feature space $Z = E_{\theta_{s}}(X) \in \mathbb{R}^{C \times H' \times W'}$, where $C$ is channel dimension. Task-specific decoders $\{H_{\theta_t}(\cdot)\}_{t=1}^T$ then transform shared representation $Z$ to diverse predictions $\{\hat{Y}_t\}_{t=1}^T$, where $\hat{Y}_t = H_{\theta_t} (Z)$. For each task $\mathcal{T}_t$, we derive the empirical loss function $\mathcal{L}_t(\theta_{s}, \theta_t)$. The conventional MTL learning objective can thus be formulated as:
\begin{equation}
\theta^* = \arg\min_\theta \sum_{t=1}^T \omega_t\mathcal{L}_t(\theta_{s}, \theta_t)
\label{eq:mtlloss}
\end{equation}
where $\omega_t > 0$ denotes task-specific weights adjusted by loss scaling methods. To prevent negative transfer, where optimizing one task deteriorates others due to the conflicting parameter updates, gradient manipulation policies harmonize parameter-wise gradients $\{\nabla_{\theta} \mathcal{L}_t\}^T_{t=1}$, especially those \wrt shared parameters $\theta_{s}$ during back-propagation:
\begin{equation}
g_t = \nabla_{\theta_{s}} \mathcal{L}_t (\theta_{s}, \theta_t), \quad
\tilde{g} = h(\{g_t\}_{t=1}^T)
\label{eq:grad_manip}
\end{equation}
where $g_t$ represents task-specific gradients and $\tilde{g}$ directly guides the updates of  $\theta_{s}$ through specifically tailored transformation function $h(\cdot)$ in optimizers: $\theta^*_{s} = \theta_{s} - \eta \cdot \tilde{g}$.

As aforementioned, this work follow the representation-centric approach in MTO to first characterize \textit{how different tasks $\{\mathcal{T}_t\}^T_{t=1}$ interact within the shared  representation space} through task saliency $\mathcal{S}_{t}$, which could be quantified by its gradient \wrt shared representation $Z$ as:
\begin{equation}
\mathcal{S}_{t} = \nabla_{Z} \mathcal{L}_t(\theta_{s}, \theta_t) \in \mathbb{R}^{B \times C \times H' \times W'}
\label{eq:rep_grad}
\end{equation}
where $B$ denotes the training batch size. The feature-level $\{\mathcal{S}_{t}\}^T_{t=1}$ here quantifies how sensitively each task's objective responds to the changes within representations~\citep{selvaraju2017GradCAM, mao2024FeatureDisentangle}. Unlike parameter gradients for direct model updates, these saliencies serve as dynamic indicators of representation-level task dependencies, providing informative learning signals for identifying and even modulating task interactions.

More importantly, this representation-centric perspective enables us to proactively approach cross-task information sharing. By regulating the task saliency, we can identify features crucial for specific tasks (helping preserve task-specific patterns and thus mitigate negative transfer) and which features show consistent importance (facilitating complementarity). Note that prior work in MTO have mainly focused on resolving conflicts while neglecting the potential for \textit{explicitly} leveraging inter-task complementarity, thus leaving this part to MTL architectures~\citep{liu2019DWA, MisraSGH16CrossStitch, yang2022CrossDistil}. 

In the following sections, we present two complementary modules based on the above task saliency $\{\mathcal{S}_t\}^T_{t=1}$: (i) an entropy-based saliency regularization to preserve task-specific learning patterns, and (ii) a sample-wise contrastive alignment to \textit{explicitly} exploit inter-task complementarity.

\subsection{Task-specific Saliency Regulation}
\label{subsec:saliency}

As aforementioned, the challenge lies in maintaining each task's effective training to prevent deterioration from negative transfer. To achieve this goal, we employ an entropy-based saliency regularization that preserves the task-specific learning patterns in representation space. As shown in Figure~\ref{fig:framework}, instead of computing task similarity like most MTO methods, we begin by aggregating channel-wise saliencies in Eq.~\ref{eq:rep_grad} to capture spatial patterns of task importance: 
\begin{equation}
\hat{\mathcal{S}}_t = \frac{1}{|C|}\sum_{c} \mathcal{S}_{t, b, c, h, w} \in \mathbb{R}^{B \times H' \times W'}
\label{eq:channel_saliency}
\end{equation}
where $\hat{\mathcal{S}}_t$ maintains the original spatial structure while reducing dimensionality for computational efficiency. Thus, $\hat{\mathcal{S}}_t$ captures how each spatial region $Z_{h,w}$ contributes to diverse task-specific learning processes. To further characterize these regions' relative importance across tasks, we normalize each $\hat{\mathcal{S}}_{i, t} \in \mathbb{R}^{B}$ into task-wise distribution $\mathcal{P}_{i, t}$:
\begin{equation}
\mathcal{P}_{i, t} = \frac{|\hat{\mathcal{S}}_{i, t}|}{\sum_{k=1}^T |\hat{\mathcal{S}}_{i, k}|}
\label{eq:ent_prob}
\end{equation}

This transforms raw spatial saliencies $\{\hat{\mathcal{S}}_{i, t}\}^T_{t=1}$ into interpretable probability distribution that identifies regions where specific tasks exhibit salient learning patterns, bridging  task- and feature-level characteristics. Holistically, high probabilities indicate regions important to individual tasks, while uniform ones suggest rather common elements. Thus, to encourage an MTL model that keeps task-salient patterns during training, we introduce an entropy-based regulation:
\begin{equation}
\mathcal{L}_{tsr} (Z) = \frac{1}{BH'W'} \sum^{BH'W'}_{i=1} (-\sum^T_{t=1} \mathcal{P}_{i, t} \log \mathcal{P}_{i, t})
\label{eq:ent_loss}
\end{equation}

This term encourages distinct task saliencies by penalizing high-entropy distributions. When spatial regions are task-salient (indicated by low entropy), the regularization $\mathcal{L}_{ts}$ encourages their task-specific patterns by penalizing excessive sharing caused by negative transfer. Empirical analysis in
Sec.~\ref{subsection:exp_ple_analysis} and Figure~\ref{fig:alpha_heads} further demonstrate the improved task-specific learning quality and successful negative transfer mitigation of $\mathcal{L}_{ts}$. The rest of this section expands on the sample-wise contrastive alignment to further promote beneficial information sharing across tasks.

\begin{table*}[t]
\centering
\caption{Performance on NYUv2~\citep{silberman2012NYUv2} dataset ($3$ indoor scene understanding tasks) with \text{DeepLabV3+}~\citep{chen2018DeepLabV3Plus} network architecture. $\uparrow (\downarrow)$ indicates higher (lower) metric values are better. The best and second-base results for each metric are highlighted in \textbf{bold} and \underline{underline}, respectively. 
$^\ddag$ indicates results from our implementation using \texttt{LibMTL}~\citep{lin2023LibMTL} codebase, which provides official support for related methods.}
\label{tab:mtl-nyu_hps}
\vspace{-0.70em}
\resizebox{\textwidth}{!}{
\begin{tabular}{lccccccccccc}
\toprule
                    & \multicolumn{2}{c}{\textbf{Segmentation}}                & \multicolumn{2}{c}{\textbf{Depth Estimation}}               & \multicolumn{5}{c}{\textbf{Surface Normal Prediction}}                                                                                & \multicolumn{1}{l}{}                 &                          \\ \cmidrule{2-10}
                    & \multicolumn{1}{c}{}     &                               &                              &                              & \multicolumn{2}{c}{Angle Dist.}                       & \multicolumn{3}{c}{Within $t^{\circ}$}                                                    & \multicolumn{1}{c}{{\bm{$\Delta{\mathrm{p}}_{\text{task}}$}${\uparrow}$}}           & {\bm{$\Delta{\mathrm{p}}_{\text{metric}}$}${\uparrow}$}                \\ \cmidrule{6-10}
Method              & \multicolumn{1}{c}{mIoU$\uparrow$} & \multicolumn{1}{c}{Pix. Acc.$\uparrow$} & \multicolumn{1}{c}{Abs. Err$\downarrow$} & \multicolumn{1}{c}{Rel. Err$\downarrow$} & \multicolumn{1}{c}{Mean$\downarrow$} & \multicolumn{1}{c}{Median$\downarrow$} & \multicolumn{1}{c}{11.25$\uparrow$} & \multicolumn{1}{c}{22.5$\uparrow$} & \multicolumn{1}{c}{30$\uparrow$} &  &        \\ \midrule
Single-Task Baseline & $53.50$ & $75.39$ & $0.3926$ & $0.1605$ & $21.99$ & $\textbf{15.16}$ & $\textbf{39.04}$ & $\textbf{65.00}$ & $\underline{75.16}$ & $0.00$                                 & $0.00$ \\ \midrule
\grow EW & $53.93$ & $75.53$ & $0.3825$ & $0.1577$ & $23.57$ & $17.01$ & $35.04$ & $60.99$ & $72.05$ & $\textcolor{resultblue}{-1.78}_{\pm0.45}$  &  $\textcolor{resultblue}{-3.85}$                    \\
GLS~\citep{chennupati2019GLS} & $\underline{54.59}$ & $\textbf{76.06}$ & $0.3785$ & $0.1555$ & $22.71$ & $16.07$ & $36.89$ & $63.11$ & $73.81$ & $\textcolor{purple}{+0.30}_{\pm0.30}$   &  $\textcolor{resultblue}{-1.10}$                        \\
RLW~\citep{lin22RLW} & $54.04$ & $75.58$ & $0.3827$ & $0.1588$ & $23.07$ & $16.49$ & $36.12$ & $62.08$ & $72.94$ & $\textcolor{resultblue}{-1.10}_{\pm0.40}$            &   $\textcolor{resultblue}{-2.64}$                       \\
UW~\citep{kendall2018UW} & $54.29$ & $75.64$ & $0.3815$ & $0.1583$ & $23.48$ & $16.92$ & $35.26$ & $61.17$ & $72.21$ & $\textcolor{resultblue}{-1.52}_{\pm0.39}$   &    $\textcolor{resultblue}{-3.54}$                 \\
DWA~\citep{liu2019DWA} & $54.06$ & $75.64$ & $0.3820$ & $0.1564$ & $23.70$ & $17.11$ & $34.90$ & $60.74$ & $71.81$ & $\textcolor{resultblue}{-1.71}_{\pm0.25}$                          &   $\textcolor{resultblue}{-3.96}$                      \\
IMTL-L~\citep{liu2021IMTL} & $53.89$ & $75.54$ & $0.3834$ & $0.1591$ & $23.54$ & $16.98$ & $35.09$ & $61.06$ & $72.12$ & $\textcolor{resultblue}{-1.92}_{\pm0.25}$       &   $\textcolor{resultblue}{-3.90}$                       \\
IGBv2~\citep{dai2023IGBv2} & $\textbf{54.61}$  & $76.00$  & $0.3817$     & $0.1576$   & $22.68$    & $15.98$     & $37.14$   & $63.25$    & $73.87$                  & $\textcolor{purple}{+0.05}_{\pm0.29}$      &      $\textcolor{resultblue}{-1.15}$                    \\
\midrule
MGDA~\citep{desideri12MGDA}  & $53.52$ & $74.76$ & $0.3852$ & $0.1566$ & $22.74$ & $16.00$ & $37.12$ & $63.22$ & $73.84$ & $\textcolor{resultblue}{-0.64}_{\pm0.25}$   &    $\textcolor{resultblue}{-1.65}$           \\
GradNorm~\citep{chen2018GradNorm} & $53.91$ & $75.38$ & $0.3842$ & $0.1571$ & $23.17$ & $16.62$ & $35.80$ & $61.90$ & $72.84$ & $\textcolor{resultblue}{-1.24}_{\pm0.15}$  &  $\textcolor{resultblue}{-2.90}$                  \\
PCGrad~\citep{yu2020PCGrad} & $53.94$ & $75.62$ & $0.3804$ & $0.1578$ & $23.52$ & $16.93$ & $35.19$ & $61.17$ & $72.19$ & $\textcolor{resultblue}{-1.57}_{\pm0.44}$    &  $\textcolor{resultblue}{-3.60}$                \\
GradDrop~\citep{chen2020GradDrop} & $53.73$ & $75.54$ & $0.3837$ & $0.1580$ & $23.54$ & $16.96$ & $35.17$ & $61.06$ & $72.07$ & $\textcolor{resultblue}{-1.85}_{\pm0.39}$    &  $\textcolor{resultblue}{-3.84}$              \\
GradVac~\citep{wang2021GradVac} & $54.21$ & $75.67$ & $0.3859$ & $0.1583$ & $23.58$ & $16.91$ & $35.34$ & $61.15$ & $72.10$ & $\textcolor{resultblue}{-1.75}_{\pm0.39}$  & $\textcolor{resultblue}{-3.72}$         \\ 
IMTL-G~\citep{liu2021IMTL} & $53.01$ & $75.04$ & $0.3888$ & $0.1603$ & $23.08$ & $16.43$ & $36.24$ & $62.23$ & $73.06$ & $\textcolor{resultblue}{-1.89}_{\pm0.54}$     & $\textcolor{resultblue}{-3.09}$     \\
CAGrad~\citep{liu2021CAGrad}  & $53.97$ & $75.54$ & $0.3885$ & $0.1588$ & $22.47$ & $15.71$ & $37.77$ & $63.82$ & $74.30$ & $\textcolor{resultblue}{-0.27}_{\pm0.35}$    &    $\textcolor{resultblue}{-0.98}$    \\
MTAdam~\citep{malkiel2021MTAdam} & $52.67$ & $74.86$ & $0.3873$ & $0.1583$ & $23.26$ & $16.55$ & $36.00$ & $61.92$ & $72.74$ & $\textcolor{resultblue}{-1.97}_{\pm0.23}$   &      $\textcolor{resultblue}{-3.36}$ \\
Nash-MTL~\citep{navon2022NashMTL} & $53.41$ & $74.95$ & $0.3867$ & $0.1612$ & $22.57$ & $15.94$ & $37.30$ & $63.40$ & $74.09$ & $\textcolor{resultblue}{-1.01}_{\pm0.13}$   & $\textcolor{resultblue}{-1.76}$       \\
MetaBalance~\citep{he2022MetaBalance} & $53.92$ & $75.57$ & $0.3901$ & $0.1594$ & $22.85$ & $16.16$ & $36.72$ & $62.91$ & $73.62$ & $\textcolor{resultblue}{-1.06}_{\pm0.17}$     &      $\textcolor{resultblue}{-2.15}$      \\
MoCo~\citep{fernando2023MoCo} & $52.25$ & $74.56$ & $0.3920$ & $0.1622$ & $22.82$ & $16.24$ & $36.58$ & $62.72$ & $73.49$ & $\textcolor{resultblue}{-2.25}_{\pm0.51}$    &  $\textcolor{resultblue}{-3.03}$     \\
Aligned-MTL~\citep{senushkin2023AlignedMTL} & $52.94$ & $75.00$ & $0.3884$ & $0.1570$ & $22.65$ & $16.07$ & $36.88$ & $63.18$ & $73.94$ & $\textcolor{resultblue}{-0.98}_{\pm0.56}$   &  $\textcolor{resultblue}{-1.92}$                        \\
FairGrad$^\ddag$~\citep{icml24FairGrad} &$53.01$  &$75.14$  &$0.3795$  & $0.1573$ & $22.51$ &$16.02$  &$36.93$  & $63.39$ & $74.17$ & $\textcolor{resultblue}{-0.47}_{\pm0.56}$ & $\textcolor{resultblue}{-1.46}$   \\
STCH$^\ddag$~\citep{icml24STCH} &$53.86$  &$75.49$  &$\underline{0.3759}$  & $\underline{0.1547}$ & $22.69$ &$16.17$  &$36.70$  & $62.96$ & $73.82$ & $\textcolor{purple}{+0.06}_{\pm0.11}$   & $\textcolor{resultblue}{-1.35}$   \\
\midrule
IMTL~\citep{liu2021IMTL} & $53.63$ & $75.44$ & $0.3868$ & $0.1592$ & $22.58$ & $15.85$ & $37.44$ & $63.52$ & $74.09$ & $\textcolor{resultblue}{-0.57}_{\pm0.24}$  & $\textcolor{resultblue}{-1.38}$   \\
DB-MTL~\citep{lin2023DBMTL} & $53.92$ & $75.60$ & $0.3768$ & $0.1557$ & $\underline{21.97}$ & $15.37$ & $\underline{38.43}$ & $\underline{64.81}$ & $\textbf{75.24}$ & $\textcolor{purple}{\underline{+1.15}}_{\pm0.16}$     &            $\textcolor{purple}{\underline{+0.56}}$              \\ 
\midrule
\brow Rep-MTL (EW)  & $\underline{54.59}$     & $\underline{76.04}$    &$\textbf{0.3750}$     &$\textbf{0.1542}$     &$\textbf{21.91}$     & $\underline{15.28}$    &  $38.37$  &  $64.72$    &  $75.05$   &       $\textcolor{purple}{\textbf{+1.70}}_{\pm0.29}$           &       $\textcolor{purple}{\textbf{+0.95}}$                  \\
\bottomrule
\end{tabular}
}
\vspace{-0.40em}
\end{table*}

\subsection{Sample-wise Cross-Task Saliency Alignment}
\label{subsec:alignment}

While saliency distribution helps preserve task-specific dynamics, hitting optimal MTL performance requires leveraging common patterns across tasks.  To address this complementary part, we present a contrastive alignment mechanism, as shown in Figure~\ref{fig:framework}, that facilitates beneficial information sharing while maintaining task distinctiveness.

The core idea behind this is that similar patterns within task saliencies $\{\mathcal{S}_{t}\}^T_{t=1}$ indicate task-generic information that could benefit multiple tasks and should be consistently represented. To identify these shared patterns, we compute the affinity maps of saliency $\mathcal{S}_{t}$ for task $\mathcal{T}_t$:
\begin{equation}
\mathcal{M}_t = \mathcal{S}_t\mathcal{S}_t^{\top} \in \mathbb{R}^{B \times C \times C},
\end{equation}
where $\mathcal{S}_t$ indicates the saliency in Eq.~\ref{eq:rep_grad}. These affinity maps capture the mutual influence patterns between features in each task's optimization, encoding how different channels interact during training. Drawing inspiration from contrastive learning~\citep{he2020MoCo, chen2020SimCLR, grill2020BYOL, chen2020SimCLRv2}, we compute the reference anchor $\mathcal{A}_b$ for each sample $b \in [B]$, which serves as candidates for information sharing in subsequent alignment:
\begin{equation}
\mathcal{A}_b = \frac{1}{H'W'}\sum_{h w} \mathcal{S}_{h w, b}, \quad \hat{\mathcal{A}_b} = \mathcal{A}_b \mathcal{A}_b^{\top}
 \in \mathbb{R}^{C \times C}
\end{equation}
where $\mathcal{S}_{hw,b}$ denotes saliency maps for sample $b$. These anchors serve as stable points for cross-task alignment, capturing potential shared dynamics that emerge across diverse tasks for identical samples. We then $L_2$-norm both anchors $\hat{\mathcal{A}_b}$ and affinities $\mathcal{M}_t$ to obtain $z^{a}_b$ and $z^t_b$, ensuring scale-invariant comparisons. For anchors $z^a_b$, we treat the related task affinities $z^t_b$ from identical sample as positive pairs, while those of different samples within a batch serve as negative pairs. As such, cross-task alignment is formulated as:
\begin{equation}
\mathcal{L}_{csa} = \frac{1}{B}\sum_{b=1}^B - \log \frac{\exp(\text{sim}(z^a_b, z^t_b) / \tau)}{\sum_{k \neq b}\exp(\text{sim}(z^a_b, z^a_k) / \tau)},
\label{eq:contrastive_loss}
\end{equation}
where $\text{sim}(\cdot,\cdot)$ denotes cosine similarity and $\tau$ controls the concentration of positive pairs $(z^a_b, z^t_b)$ relative to the negative ones $(z^a_b, z^a_k)$. As such, the balance between positive and negative pairs as well as the inclusion of batch information maintains task distinctiveness while promoting information sharing across tasks. Empirical analysis in Sec.~\ref{subsection:exp_ple_analysis} and  Figure~\ref{fig:alpha_backbone} shows that this contrastive saliency alignment facilitates models' optimization towards the overall MTL objectives, indicating that the intrinsic inter-task complementarity has been effectively excavated by CSA $\mathcal{L}_{csa}$.

\subsection{Joint Optimization}

To achieve robust multi-task training that both preserves task-specific learning signals and enables cross-task sharing, we combine MTL objectives with above regularization:
\begin{equation}
\mathcal{L}_\text{Rep}=\sum^{T}_{t=1} \mathcal{L}_{t} (\theta_{s}, \theta_t) + \lambda_{tsr} \mathcal{L}_{tsr} (Z) + \lambda_{csa} \mathcal{L}_{csa} (Z),
    \label{eq:total_loss}
\end{equation}
where $\lambda_{1}$ and $\lambda_{2}$ balance the influence of above two regularization terms. During joint optimization, gradients from all these components flow through the multi-task model, introducing implicit adjustment to model parameter updates.
\section{Experiment}

\begin{table}[t]
\centering
\caption{Performance on Cityscapes~\citep{cvpr16Cityscapes} dataset ($2$ scene understanding tasks) with \text{DeepLabV3+}~\citep{chen2018DeepLabV3Plus} architecture. $\uparrow (\downarrow)$ indicates higher (lower) metric values are better. The best and second-best results are marked in \textbf{bold} and \underline{underline}, respectively. $^\ddag$ indicates results from our implementation using \texttt{LibMTL}~\citep{lin2023LibMTL} codebase.}
\label{tab:mtl-city_hps}
\vspace{-0.70em}
\resizebox{1.00\linewidth}{!}{
\begin{tabular}{lccccc}
\toprule
               & \multicolumn{2}{c}{\textbf{Segmentation}} & \multicolumn{2}{c}{\textbf{Depth Estimation}} & \multicolumn{1}{c}{{\bm{$\Delta{\mathrm{p}}_{\text{task}}$}${\uparrow}$}}                                                                                          \\ \cmidrule{2-5}
Method         & \multicolumn{1}{c}{mIoU$\uparrow$}          & \multicolumn{1}{c}{Pix. Acc.$\uparrow$}        & \multicolumn{1}{c}{Abs. Err$\downarrow$}          & \multicolumn{1}{c}{Rel. Err$\downarrow$}      &         \\ \midrule
Single-Task Baseline & $69.06$ & $91.54$ & $0.01282$ & $43.53$ & $0.00$                                                                             \\ \midrule
\grow EW & $68.93$ & $91.58$ & $0.01315$ & $45.90$ & $\textcolor{resultblue}{-2.05}_{\pm0.56}$   \\
GLS~\citep{chennupati2019GLS} & $68.69$ & $91.45$ & $0.01280$ & $44.13$ & $\textcolor{resultblue}{-0.39}_{\pm1.06}$    \\
RLW~\citep{lin22RLW} & $69.03$ & $91.57$ & $0.01343$ & $44.77$ & $\textcolor{resultblue}{-1.91}_{\pm0.21}$   \\
UW~\citep{kendall2018UW} & $69.03$ & $91.61$ & $0.01338$ & $45.89$ & $\textcolor{resultblue}{-2.45}_{\pm0.68}$     \\
DWA~\citep{liu2019DWA} & $68.97$ & $91.58$ & $0.01350$ & $45.10$ & $\textcolor{resultblue}{-2.24}_{\pm0.28}$    \\
IMTL-L~\citep{liu2021IMTL} & $68.98$ & $91.59$ & $0.01340$ & $45.32$ & $\textcolor{resultblue}{-2.15}_{\pm0.88}$       \\
IGBv2~\citep{dai2023IGBv2} & $68.44$ & $91.31$ & $0.01290$           & $45.03$ & $\textcolor{resultblue}{-1.31}_{\pm0.61}$    \\
\midrule
MGDA~\citep{desideri12MGDA} & $69.05$ & $91.53$ & ${0.01280}$ & $44.07$ & $\textcolor{resultblue}{-0.19}_{\pm0.30}$     \\
GradNorm~\citep{chen2018GradNorm} & $68.97$ & $91.60$ & $0.01320$ & $44.88$ & $\textcolor{resultblue}{-1.55}_{\pm0.70}$       \\
PCGrad~\citep{yu2020PCGrad}  & $68.95$ & $91.58$ & $0.01342$ & $45.54$ & $\textcolor{resultblue}{-2.36}_{\pm1.17}$  \\
GradDrop~\citep{chen2020GradDrop} & $68.85$ & $91.54$ & $0.01354$ & $44.49$ & $\textcolor{resultblue}{-2.02}_{\pm0.74}$   \\
GradVac~\citep{wang2021GradVac} & $68.98$ & $91.58$ & $0.01322$ & $46.43$ & $\textcolor{resultblue}{-2.45}_{\pm0.54}$      \\
IMTL-G~\citep{liu2021IMTL}  & $69.04$ & $91.54$ & $0.01280$ & $44.30$ & $\textcolor{resultblue}{-0.46}_{\pm0.67}$        \\
CAGrad~\citep{liu2021CAGrad}  & $68.95$ & $91.60$ & $0.01281$ & $45.04$ & $\textcolor{resultblue}{-0.87}_{\pm0.88}$     \\
MTAdam~\citep{malkiel2021MTAdam}  & $68.43$ & $91.26$ & $0.01340$ & $45.62$ & $\textcolor{resultblue}{-2.74}_{\pm0.20}$       \\
Nash-MTL~\citep{navon2022NashMTL} & $68.88$ & $91.52$ & $\textbf{0.01265}$ & $45.92$ & $\textcolor{resultblue}{-1.11}_{\pm0.21}$    \\
MetaBalance~\citep{he2022MetaBalance} & $69.02$ & $91.56$ & $\underline{0.01270}$ & $45.91$ & $\textcolor{resultblue}{-1.18}_{\pm0.58}$      \\
MoCo~\citep{fernando2023MoCo} & $\underline{69.62}$ & $\underline{91.76}$ & $0.01360$ & $45.50$ & $\textcolor{resultblue}{-2.40}_{\pm1.50}$     \\
Aligned-MTL~\citep{senushkin2023AlignedMTL} & $69.00$ & $91.59$ & $\underline{0.01270}$ & $44.54$ & $\textcolor{resultblue}{-0.43}_{\pm0.44}$       \\
FairGrad$^\ddag$~\citep{icml24FairGrad} & $68.84$ &$91.48$  &$\underline{0.01270}$  & $46.26$  &$\textcolor{resultblue}{-1.43}_{\pm0.63}$    \\
STCH$^\ddag$~\citep{icml24STCH} & $68.21$ & $91.23$ & $0.01280$ & $\textbf{43.17}$ & $\textcolor{resultblue}{-0.15}_{\pm0.38}$    \\
\midrule
IMTL~\citep{liu2021IMTL}  & $69.07$ & $91.55$ & ${0.01280}$ & $44.06$ & $\textcolor{resultblue}{-0.32}_{\pm0.10}$   \\
DB-MTL~\citep{lin2023DBMTL}  & $69.17$ & $91.56$ & ${0.01280}$ & $43.46$ & $\textcolor{purple}{\underline{+0.20}}_{\pm0.40}$            \\ 
\midrule
\brow Rep-MTL (EW) &  $\textbf{69.72}$  & $\textbf{91.85}$    &  $\underline{0.01270}$    &  $\underline{43.42}$       &    $\textcolor{purple}{\textbf{+0.62}}_{\pm0.53}$                                                           \\ \bottomrule
\end{tabular}}
\vspace{-0.40em}
\end{table}

We evaluate Rep-MTL on four MTO benchmarks covering both task-shift and domain-shift scenarios: (i) NYUv2~\citep{silberman2012NYUv2} ($3$-task indoor scene understanding), (ii) Cityscapes~\citep{cvpr16Cityscapes} ($2$-task urban scene understanding), (iii) Office-31~\citep{saenko2010Office31} ($3$-domain image classification), and (iv) Office-Home~\citep{venkateswara2017OfficeHome} ($4$-domain image classification). To ensure thorough and rigorous evaluation, we follow the more challenging benchmark settings~\citep{lin2023DBMTL} for NYUv2 and Cityscapes, where most baselines exhibit negative performance gains (see Table~\ref{tab:mtl-nyu_hps}, ~\ref{tab:mtl-city_hps}).

In this section, we first introduce the baselines and evaluation metrics, followed by quantitative results on scene understanding (Sec.~\ref{subsec:exp_sc}) and image classification (Sec.~\ref{subsec:exp_ic}). We then present PL exponent analysis (Sec.~\ref{subsection:exp_ple_analysis}) with ablation studies (Sec.~\ref{subsec:exp_ablation}) that validate Rep-MTL's effectiveness beyond benchmarking results. There are also evaluations to assess Rep-MTL's robustness and applicability, including sensitivity of hyperparameters (Sec.~\ref{subsec:exp_hyper}) and learning rates (Appendix~\ref{app:learning_rate}), and also its optimization speed (Sec.~\ref{subsec:exp_runtime}).

\vspace{-1.00em}
\paragraph{Baselines.}
We compare our Rep-MTL against 23 popular MTO algorithms, including loss scaling policies (GLS~\citep{chennupati2019GLS},
RLW~\citep{lin22RLW},
UW~\citep{kendall2018UW},
DWA~\citep{liu2019DWA},
IGBv2~\citep{dai2023IGBv2}), gradient manipulation (MGDA~\citep{desideri12MGDA},
GradNorm~\citep{chen2018GradNorm},
PCGrad~\citep{yu2020PCGrad},
GradDrop~\citep{chen2020GradDrop},
GradVac~\citep{wang2021GradVac},
CAGrad~\citep{liu2021CAGrad}, MTAdam~\citep{malkiel2021MTAdam},
Nash-MTL~\citep{navon2022NashMTL},
MetaBalance~\citep{he2022MetaBalance},
MoCo~\citep{fernando2023MoCo}, Aligned-MTL~\citep{senushkin2023AlignedMTL}), and hybrid approaches (IMTL~\citep{liu2021IMTL}, DB-MTL~\citep{lin2023DBMTL}). We also implement recent FairGrad~\citep{icml24FairGrad}, and STCH~\citep{icml24STCH} on NYUv2~\citep{silberman2012NYUv2} and Cityscapes~\citep{cvpr16Cityscapes} using the open-source \texttt{LibMTL}~\citep{lin2023LibMTL} codebase. For all included algorithms, HPS architecture is employed for fair comparison.

\vspace{-1.00em}
\paragraph{Evaluation Metrics.}
Task evaluation follows established metrics~\citep{liu2019DWA, senushkin2023AlignedMTL, lin2023DBMTL}: For semantic segmentation, we use mean Intersection over Union (mIoU) and pixel-wise accuracy (Pix Acc). Depth estimation performance is measured using relative error (Rel Err) and absolute error (Abs Err). Surface normal prediction is evaluated with mean and median angle errors, along with percentage of normals within angular thresholds of $t^{\circ}~(t=11.25, 22.5, 30)$. We quantify MTL performance improvements relative to STL baselines as average gains over tasks $\Delta{\mathrm{p}}_\text{task}$ and metrics $\Delta{\mathrm{p}}_\text{metric}$: 
\begin{align}
\Delta{\mathrm{p}}_\text{metric} =\frac{1}{T} \sum_{t=1}^{T}(-1)^{\sigma_{t}} \frac{\left(M_{m, t}-M_{b, t}\right)} {M_{b, t}} , \\ 
\Delta{\mathrm{p}}_\text{task}=\frac{1}{T} \sum_{t=1}^{T} \sum_{k=1}^{n_t}(-1)^{\sigma_{tk}} \frac{\left(M_{m, tk}-M_{b, tk}\right) } {M_{b, tk}}
\label{eq:delta_p}
\end{align}
where $T$ denotes the total number of tasks, $n_t$ represents the number of evaluation metrics for task $t$, and $M_{m,t}$ indicates the performance of method $m$ on task $t$. The sign coefficient $\sigma_{t}$ ($\sigma_{tk}$) equals $0$ for positive metrics (\eg, mIoU, Pix Acc) and $1$ for negative metrics (\eg, Err, Dist.), ensuring consistent interpretation. Each experiment is repeated three times. Please refer to Appendix~\ref{app:implement} for experimental settings.

\subsection{Scene Understanding Tasks}
\label{subsec:exp_sc}
\paragraph{Datasets and Settings.}
We evaluate our Rep-MTL on two scene understanding benchmarks: (i) NYUv2~\citep{silberman2012NYUv2} comprises three tasks ($13$-class semantic segmentation, depth estimation, and surface normal prediction), with $795$ training and $654$ validation images; (ii) Cityscapes~\citep{cvpr16Cityscapes} owns two tasks ($7$-class semantic segmentation and depth estimation) with $2975$ training and $500$ testing images. Following previous settings~\citep{lin2023DBMTL}, we employ DeepLabV3+~\citep{chen2018DeepLabV3Plus} as our architecture, where a dilated ResNet-50~\citep{he2016ResNet} as the encoder and ASPP as decoders. Please view Appendix~\ref{app:implement} for details.

\vspace{-1.00em}
\paragraph{NYUv2 Results.}
Table~\ref{tab:mtl-nyu_hps} shows that Rep-MTL achieves competitive multi-task performance gains on the NYUv2 benchmark, as measured by $\Delta{\mathrm{p}}_{\text{task}}$ and $\Delta{\mathrm{p}}_{\text{metric}}$. \textbf{(i) EW baseline comparison:} Compared to the gray-marked EW baseline, Rep-MTL shows significant improvements, with gains of $+3.48$ in $\Delta{\mathrm{p}}_{\text{task}}$ and $+4.8$ in $\Delta{\mathrm{p}}_{\text{metric}}$, which improves performance the most without extra optimizer modifications. \textbf{(ii) SOTA comparison:} Rep-MTL outperforms previous leading methods, DB-MTL~\citep{lin2023DBMTL}, by about $48\%$ in $\Delta{\mathrm{p}}_{\text{task}}$ ($+1.70$ \vs $+1.15$) and nearly $\sim70\%$ in $\Delta{\mathrm{p}}_{\text{metric}}$ ($+0.95$ \vs $+0.56$). Furthermore, Rep-MTL surpasses DB-MTL in $6/9$ sub-task metrics, showcasing its effectiveness.

\vspace{-1.00em}
\paragraph{Cityscapes Results.}
Table~\ref{tab:mtl-city_hps} demonstrates Rep-MTL's effectiveness on the Cityscapes benchmark, where it achieves the best results in semantic segmentation and the second-best results in two depth estimation metrics. \textbf{(i) EW baseline comparison:} Rep-MTL improves upon EW baseline by $+2.67$ in $\Delta{\mathrm{p}}_{\text{task}}$, showcasing its ability to improve MTL performance across datasets. \textbf{(ii) SOTA comparison:} DB-MTL~\citep{lin2023DBMTL} is the only method that surpasses STL baseline on this challenging benchmark setting. However, Rep-MTL slightly exceeds the powerful DB-MTL in $\Delta{\mathrm{p}}_{\text{task}}$ ($+0.62$ \vs $+0.20$), which indicates an improvement of multi-task dense prediction in outdoor scene understanding scenarios.

\subsection{Image Classification Tasks}
\label{subsec:exp_ic}
\paragraph{Datasets and Settings.}
To further validate Rep-MTL's effectiveness in domain-shift scenarios, the following datasets are included: (i) Office-31~\citep{saenko2010Office31} which contains $4110$ images from three domains (tasks): Amazon, DSLR, and Webcam. Each task has $31$ object categories. (ii) Office-Home~\citep{venkateswara2017OfficeHome} which contains $15500$ images from four domains (tasks): artistic, clipart, product, and real-world. Each of them has $65$ object categories in office and home settings. We use the data split as $60\%$ for training, $20\%$ for validation, and $20\%$ for testing. Following \citet{lin2023DBMTL}, we use ResNet18 as network architecture. Please view Appendix~\ref{app:implement} for details.

\vspace{-1.00em}
\paragraph{Results.}
As shown in Table~\ref{tab:mtl-office-home}, Rep-MTL still maintains its effectiveness in domain-shift scenarios~\citep{venkateswara2017OfficeHome}. Concretely, Rep-MTL surpasses EW by $+0.97$ in average performance and achieves a positive $\Delta{\mathrm{p}_\text{task}}$ of $+0.41$, which advances the previous SOTA by approximately $140\%$ ($+0.41$ \vs $+0.17$). This validates Rep-MTL's capability to exploit inter-task complementarity even if there are domain gaps.
Situations on Office-31~\citep{saenko2010Office31} (Appendix~\ref{app:results}) are even more challenging, where Rep-MTL exhibits the best results on Webcam, average performance (Avg.), and $\Delta{\mathrm{p}}_{\text{task}}$ improvement, slightly outperforming previous SOTA by $25\%$ ($+1.31$ \vs $+1.05$).

\subsection{Power Law (PL) Exponent Analysis}
\label{subsection:exp_ple_analysis}

\begin{figure}[t]
    % \vspace{-0.25em}
    \begin{center}
    \includegraphics[width=1.00\linewidth]{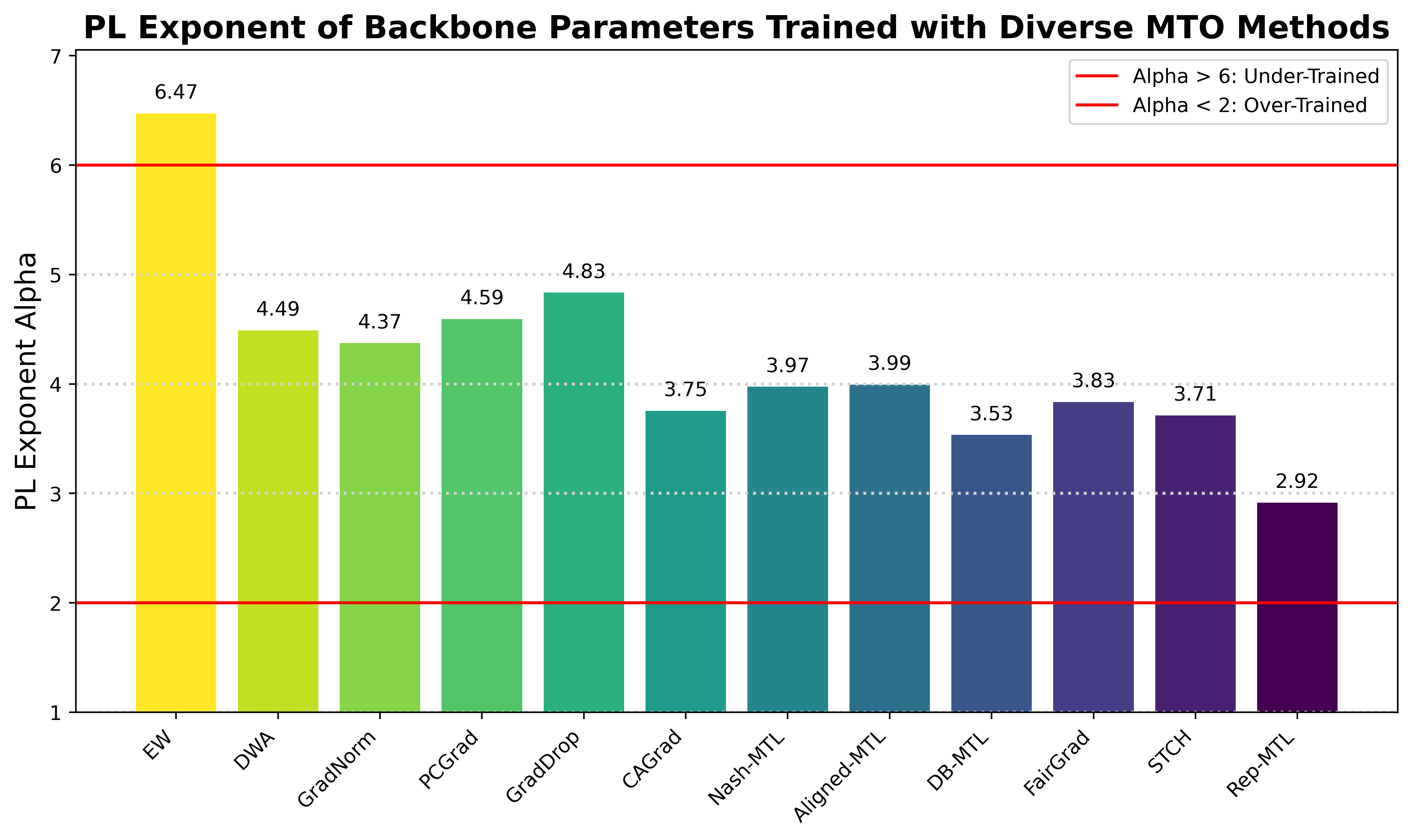}
    \hspace{-1.4em}
    \end{center}
    \vspace{-3.20em}
    \caption{Comparison of PL exponent alpha~\citep{ NC2021WeightWatcher} for backbone parameters trained with  
diverse MTO methods on NYUv2~\citep{silberman2012NYUv2}. It validates how well the backbone adapts to MTL objectives, where lower values indicate more effective training. Values outside $[2,4]$ suggest potential over- or under-training. We leverage this to show how methods affect model updates, as well-trained backbones suggest beneficial cross-task sharing to the overall MTL objectives.}
    \label{fig:alpha_backbone}
    \vspace{-0.75em}
\end{figure}

\begin{figure*}[t]
    % \vspace{-0.25em}
    \begin{center}
    \includegraphics[width=1.00\linewidth]{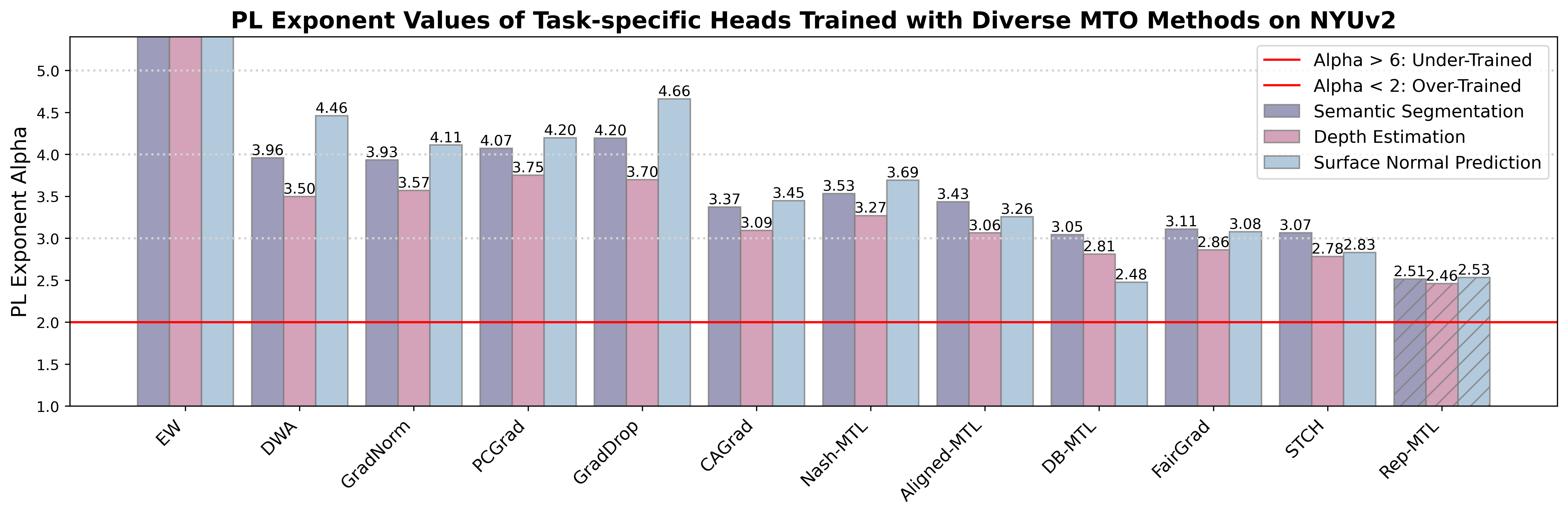}
    \hspace{-1.5em}
    \end{center}
    \vspace{-3.26em}
    \caption{Visualization of PL exponent alpha~\citep{JMLR2021selfreg, NC2021WeightWatcher} for tasks-specific heads (3 tasks) trained with  
different MTO methods on NYUv2~\citep{silberman2012NYUv2}. PL exponent quantifies how well each decoder adapts to its task-specific objective, where lower values practically indicate more effective model training. Values outside the range $[2,6]$ suggest potential over- or under-training due to task conflicts. The variation across decoders of each method indicates training imbalance. We employ this to evaluate how diverse MTO approaches influence decoder parameter updates, as well-trained decoders should exhibit both \textit{low and balanced} values, indicating effective individual task training with successful negative transfer mitigation. For display, we constrain $\alpha$ range to $(0, 6)$ though EW results ($10.26$, $7.01$, $14.25$) extend beyond this range.} 
    \label{fig:alpha_heads}
    \vspace{-0.75em}
\end{figure*}

We analyze Power Law (PL) exponent alpha~\citep{JMLR2021selfreg, NC2021WeightWatcher} derived from Heavy-Tailed Self-Regularization (HT-SR) theory~\citep{martin2021implicit, mahoney2019traditionalHTSR}. For each layer's weight matrix $W$, the PL exponent is computed by fitting the Empirical Spectral Density of $X = W^TW$ to $\rho(\lambda) \sim \lambda^{-\alpha}$, where $\lambda$ are eigenvalues of the correlation matrix. This scale-invariant metric has proven effective in evaluating training quality without access to training or test data~\citep{arxiv2024BOCB, arxiv2024OpenMixup}, making it particularly suitable for analyzing MTL models: $\alpha$ quantifies how well different model parts (\eg, shared backbone or task-specific decoders) adapt to their corresponding objectives (\eg, the overall MTL loss or task-specific losses). Thus, by examining $\alpha$ values of different components of MTL model parameters, we can validate both the MTO methods' ability to assist the training process, \ie, mitigating negative transfer among tasks while exploiting inter-task complementarity.
In practice, well-trained models typically exhibit $\alpha \in [2,4]$, while poorly-trained or over-parameterized models tend to show $\alpha \gg 4$, providing an applicable tool for assessing the overall effectiveness of MTO methods. In particular, a low $\alpha$ value for backbone often indicates effective cross-task information sharing for the overall MTL objectives, while low and balanced values across heads suggest high-quality training of individual tasks, thus indicating less negative transfer.

\vspace{0.20em}
In this work, we analyze DeepLabV3+ models trained with different MTO methods on NYUv2~\citep{silberman2012NYUv2}. As shown in Figure~\ref{fig:alpha_backbone}, models trained with Rep-MTL exhibit superior $\alpha = 2.92$ for shared parameters $\theta_s$ compared to other MTO methods, which indicates more favorable inter-task complementarity. Moreover, Figure~\ref{fig:alpha_heads} reveals that Rep-MTL yields lower and more balanced PL exponent ($2.51, 2.46, 2.53$) for decoders, demonstrating effective training of individual tasks, thereby mitigating negative transfer. Please view Appendix~\ref{app:empirical} for more details. We also incorporate PL exponent analysis into ablation studies in Sec.~\ref{subsec:exp_ablation} to explore how the TSR and CSA modules separately influence model training.

\begin{table}[]
\centering
\caption{Performance on Office-Home~\citep{venkateswara2017OfficeHome} dataset with $4$ diverse image classification tasks.  $\uparrow$ indicates the higher the metric values, the better the methods' performance. The best and second-best results of each metric are marked in \textbf{bold} and \underline{underline}, respectively.}
\label{tab:mtl-office-home}
\vspace{-0.70em}
\resizebox{1.00\linewidth}{!}{
\begin{tabular}{lcccccc}
\toprule
Method               & \multicolumn{1}{c}{\textbf{Artistic}} & \multicolumn{1}{c}{\textbf{Clipart}} & \multicolumn{1}{c}{\textbf{Product}} & \multicolumn{1}{c}{\textbf{Real}} & \multicolumn{1}{c}{\textbf{Avg.}$\uparrow$} & \multicolumn{1}{c}{\bm{$\Delta{\mathrm{p}}_{\text{task}}$}${\uparrow}$}  \\ \midrule
Single-Task Baseline & $65.59$ & $\textbf{79.60}$ & $\textbf{90.47}$ & $80.00$ & $78.91$ & $0.00$ \\ \midrule
\grow EW & $65.34$ & $78.04$ & $89.80$ & $79.50$ & $78.17_{\pm0.37}$ & $\textcolor{resultblue}{-0.92}_{\pm0.59}$     \\
GLS~\citep{chennupati2019GLS} & $64.51$ & $76.85$ & $89.83$ & $79.56$ & $77.69_{\pm0.27}$ & $\textcolor{resultblue}{-1.58}_{\pm0.46}$     \\
RLW~\citep{lin22RLW} & $64.96$ & $78.19$ & $89.48$ & $\textbf{80.11}$ & $78.18_{\pm0.12}$ & $\textcolor{resultblue}{-0.92}_{\pm0.14}$    \\
UW~\citep{kendall2018UW} & $65.97$ & $77.65$ & $89.41$ & $79.28$ & $78.08_{\pm0.30}$ & $\textcolor{resultblue}{-0.98}_{\pm0.46}$  \\
DWA~\citep{liu2019DWA} & $65.27$ & $77.64$ & $89.05$ & $79.56$ & $77.88_{\pm0.28}$ & $\textcolor{resultblue}{-1.26}_{\pm0.49}$ \\
IMTL-L~\citep{liu2021IMTL} & $65.90$ & $77.28$ & $89.37$ & $79.38$ & $77.98_{\pm0.38}$ & $\textcolor{resultblue}{-1.10}_{\pm0.61}$ \\
IGBv2~\citep{dai2023IGBv2} & $65.59$ & $77.57$ & $89.79$ & $78.73$ & $77.92_{\pm0.21}$      &         $\textcolor{resultblue}{-1.21}_{\pm0.22}$  \\
\midrule
MGDA~\citep{desideri12MGDA} & $64.19$ & $77.60$ & $89.58$ & $79.31$ & $77.67_{\pm0.20}$ & $\textcolor{resultblue}{-1.61}_{\pm0.34}$   \\
GradNorm~\citep{chen2018GradNorm} & $66.28$ & $77.86$ & $88.66$ & $79.60$ & $78.10_{\pm0.63}$ & $\textcolor{resultblue}{-0.90}_{\pm0.93}$  \\
PCGrad~\citep{yu2020PCGrad} & $66.35$ & $77.18$ & $88.95$ & $79.50$ & $77.99_{\pm0.19}$ & $\textcolor{resultblue}{-1.04}_{\pm0.32}$  \\
GradDrop~\citep{chen2020GradDrop} & $63.57$ & $77.86$ & $89.23$ & $79.35$ & $77.50_{\pm0.23}$ & $\textcolor{resultblue}{-1.86}_{\pm0.24}$   \\
GradVac~\citep{wang2021GradVac} & $65.21$ & $77.43$ & $89.23$ & $78.95$ & $77.71_{\pm0.19}$ & $\textcolor{resultblue}{-1.49}_{\pm0.28}$ \\
IMTL-G~\citep{liu2021IMTL} & $64.70$ & $77.17$ & $89.61$ & $79.45$ & $77.98_{\pm0.38}$ & $\textcolor{resultblue}{-1.10}_{\pm0.61}$    \\
CAGrad~\citep{liu2021CAGrad} & $64.01$ & $77.50$ & $89.65$ & $79.53$ & $77.73_{\pm0.16}$ & $\textcolor{resultblue}{-1.50}_{\pm0.29}$  \\
MTAdam~\citep{malkiel2021MTAdam} & $62.23$ & $77.86$ & $88.73$ & $77.94$ & $76.69_{\pm0.65}$ & $\textcolor{resultblue}{-2.94}_{\pm0.85}$ \\
Nash-MTL~\citep{navon2022NashMTL} & $66.29$ & $78.76$ & $90.04$ & $\textbf{80.11}$ & $78.80_{\pm0.52}$ & $\textcolor{resultblue}{-0.08}_{\pm0.69}$ \\
MetaBalance~\citep{he2022MetaBalance} & $64.01$ & $77.50$ & $89.72$ & $79.24$ & $77.61_{\pm0.42}$ & $\textcolor{resultblue}{-1.70}_{\pm0.54}$ \\
MoCo~\citep{fernando2023MoCo} & $63.38$ & $\underline{79.41}$ & $90.25$ & $78.70$ & $77.93$ & -     \\
Aligned-MTL~\citep{senushkin2023AlignedMTL} & $64.33$ & $76.96$ & $89.87$ & $79.93$ & $77.77_{\pm0.70}$ & $\textcolor{resultblue}{-1.50}_{\pm0.89}$ \\
\midrule
IMTL~\citep{liu2021IMTL} & $64.07$ & $76.85$ & $89.65$ & $79.81$ & $77.59_{\pm0.29}$ & $\textcolor{resultblue}{-1.72}_{\pm0.45}$   \\
DB-MTL~\citep{lin2023DBMTL} & $\textbf{67.42}$ & $77.89$  & $\underline{90.43}$  & $80.07$    & $\underline{78.95}_{\pm0.35}$               & $\textcolor{purple}{\underline{+0.17}}_{\pm0.44}$      \\
\midrule
\brow Rep-MTL (EW) & $\underline{67.40}$     &     $$78.75$$     &  $90.37$                           &   $80.04$         &   $\textbf{79.14}_{\pm0.41}$      &  $\textcolor{purple}{\textbf{+0.41}}_{\pm0.58}$             \\ \bottomrule
\end{tabular}}
\vspace{-0.8em}
\end{table}

\subsection{Ablation Study}
\label{subsec:exp_ablation}
The proposed Rep-MTL includes two key components: (i) TSR optimizing individual task learning via entropy-based regularization, and (ii) CSA which facilitates inter-task information sharing. Thus, to validate the contribution of each component, we conduct ablation studies using standard performance metrics and PL exponent analysis on NYUv2~\citep{silberman2012NYUv2} and Cityscapes~\citep{cvpr16Cityscapes} with results in Table~\ref{tbl:ablation} and Appendix~\ref{app:alpha_ablation}.

\vspace{-1.0em}
\paragraph{The CSA Module.}
Table~\ref{tbl:ablation} examines four configurations: (i) a baseline using neither component; (ii) CSA only; (iii) TSR only; and (iv) complete Rep-MTL with both components. The results demonstrate that both TSR and CSA individually contribute to positive $\Delta {\mathrm{p}_{\text{task}}}$, while their combination yields the best results, showcasing their effectiveness. Notably, CSA alone yields greater performance gains than TSR alone, which is corroborated by our PL exponent analysis (Appendix~\ref{app:ple_ablation_sr}). It shows that even without TSR, CSA helps maintain balanced PL exponents across tasks-specific heads relative to baselines, suggesting that CSA effectively enables the model to leverage inter-task complementarity for joint training. This also shows the importance of \textit{explicit designs for exploiting inter-task complementarity} in MTO.

\vspace{-1.0em}
\paragraph{The TSR Module.}
Beyond standard performance metrics, PL exponent analysis provides deeper insights into how each component functions.
As shown in Appendix~\ref{app:ple_ablation_sa}, CSA leads to lower backbone PL exponent $\alpha$ within optimal range, indicating more effective task-generic feature learning from shared information. Meanwhile, TSR yields lower and more balanced PL exponents across task-specific decoders (Appendix~\ref{app:ple_ablation_sr}), demonstrating its capability in maintaining effective training of individual tasks, thereby mitigating negative transfer. It shows that \textit{tackling negative transfer could go beyond pure conflict-resolving designs}. Approaches like TSR, which aim to emphasize individual tasks' training, may represent a promising and complementary direction. Together, these findings provide compelling evidence for both TSR and CSA's effectiveness.

\subsection{Hyperparameter Sensitivity}
\label{subsec:exp_hyper}
For practical deployment, robust algorithms are expected to maintain stable performance across a reasonable range of hyperparameter settings. This directly impacts the method's applicability, especially in resource-constrained scenarios.

\begin{table}[!h]
\centering
\caption{Ablation study of Rep-MTL comprising two complementary components on NYUv2~\citep{silberman2012NYUv2} and Cityscapes~\citep{cvpr16Cityscapes} in terms of task-level performance gains $\Delta {\mathrm{p}}_{\text{task}}$${\uparrow}$ relative to STL baselines.} 
\label{tbl:ablation}
\vspace{-0.70em}
\resizebox{1.00\linewidth}{!}{
\begin{tabular}{cc|cc}
\toprule
Cross-task Saliency & Task-specific Saliency  & {NYUv2} &{Cityscapes} \\
Alignment (CSA) & Regulation (TSR) & {\bm{$\Delta{\mathrm{p}}_{\text{task}}$}${\uparrow}$} & {\bm{$\Delta{\mathrm{p}}_{\text{task}}$}${\uparrow}$}\\
\midrule
\xmarkg & \xmarkg & ${-1.78}_{\pm0.45}$ & ${-2.05}_{\pm0.56}$ \\
\grow \Checkmark & \xmarkg & ${+1.06}_{\pm0.27}$ & ${+0.21}_{\pm0.68}$ \\
\xmarkg &  \Checkmark & ${+0.23}_{\pm0.29}$ & ${-0.34}_{\pm0.24}$ \\
\grow \Checkmark &  \Checkmark & $\textbf{+1.70}_{\pm0.29}$ & $\textbf{+0.62}_{\pm0.53}$ \\
\bottomrule
\end{tabular}
}
\vspace{-0.40em}
\end{table}

Empirical analysis in Appendix~\ref{app:hyper} verifies that our Rep-MTL consistently achieves positive gains ($\Delta \mathrm{p}_{\text{task}} > 0$)  across a wide range of weighting coefficients ($\lambda_{tsr}, \lambda_{csa} \in [0.7, 1.5]$), which reduces the need for meticulous hyperparameter tuning. We also analyze Rep-MTL's learning rate sensitivity in Appendix~\ref{app:learning_rate} to further show its robustness.

\subsection{Efficiency Analysis}
\label{subsec:exp_runtime}
Optimization speed remains crucial for MTL. Thus, we conduct empirical analysis on NYUv2~\citep{silberman2012NYUv2} in Appendix~\ref{app:runtime}. while Rep-MTL requires increased runtime than loss scaling due to gradient computation, it exhibits better efficiency compared to most gradient manipulation methods approximately 26\% faster than Nash-MTL~\citep{navon2022NashMTL} and 12\% faster than FairGrad~\citep{icml24FairGrad}), while delivering superior performance gains.

\section{Conclusion}
\label{sec:conclusion}
This paper presents Rep-MTL, a regularization-based MTO approach that leverages representation-level task saliency to advance multi-task training. By operating directly on task representation space, Rep-MTL aims to preserve the effective training of individual tasks while explicitly exploiting inter-task complementarity. Experiments not only reveal Rep-MTL's competitive performance, but highlight the significant yet largely untapped potential of directly regularizing representation space for more effective MTL systems.

{
    \small
    \bibliographystyle{ieeenat_fullname}
    \bibliography{main}
}

\clearpage
%\setcounter{page}{1}
%\maketitlesupplementary

\appendix

\section*{\large{Appendix}}
This appendix offers additional empirical analyses, experimental results, and further discussions of our work. The appendix sections are organized as follows:
\begin{itemize}[leftmargin=1.0em]

\vspace{0.25em}
\item In Appendix~\ref{app:implement}, we provide experimental setups and implementation specifications across all four benchmarks in this paper, including NYUv2~\citep{silberman2012NYUv2}, Cityscapes~\citep{cvpr16Cityscapes}, Office-Home~\citep{venkateswara2017OfficeHome}, and Office-31~\citep{saenko2010Office31}. This includes comprehensive information on employed network architectures, optimization algorithms, training protocols, loss functions, and hyper-parameter configurations.

\vspace{0.25em}
\item In Appendix~\ref{app:results}, we provide complete experimental results on the Office-31 dataset, which were omitted from the main manuscript due to space constraints. We also discuss the proposed Rep-MTL method combined with all experimental results from four benchmarks.

\vspace{0.25em}
\item In Appendix~\ref{app:alpha_ablation}, we present additional ablation studies through the lens of PL exponent alpha analysis~\citep{JMLR2021selfreg, NC2021WeightWatcher}. These studies further demonstrate how each mechanism of Rep-MTL contributes to facilitating cross-task positive transfer while preserving task-specific learning patterns, thereby mitigating the negative transfer in MTL.

\vspace{0.25em}
\item In Appendix~\ref{app:empirical}, we conduct experiments to validate Rep-MTL's robustness and practical applicability, with particular emphasis on the sensitivity of hyper-parameters $\lambda_{tsr}, \lambda_{csa}$, learning rates, and its optimization speed.
\end{itemize}

\section{Implementation Details}
\label{app:implement}
This appendix section provides an expansion of the experimental configurations and implementation specifications of the experiments from the main manuscript. We detail the network architectures, optimizers, and training recipes for each included benchmark to ensure reproducibility.

\vspace{-0.28em}
\paragraph{NYUv2 Dataset}
Following the implementations in previous studies~\citep{lin22RLW, lin2023DBMTL}, we employ the DeepLabV3+~\citep{chen2018DeepLabV3Plus} network architecture, containing a dilated ResNet 50~\citep{he2016ResNet} backbone pre-trained on ImageNet and the Atrous Spatial Pyramid Pooling (ASPP) as task-specific decoders. The MTL model is trained for $200$ epochs using the Adam optimizer with an initial learning rate of $10^{-4}$ and weight decay of $10^{-5}$. Consistent with prior works~\citep{lin22RLW, lin2023DBMTL}, we implement a learning rate schedule where the rate is halved to $5\times10^{-5}$ after $100$ training epochs. For the three tasks on NYUv2~\citep{silberman2012NYUv2}, we utilize cross-entropy loss for semantic segmentation, $L_1$ loss for depth estimation, and cosine loss for surface normal prediction. We adopted the same logarithmic transformation as in previous works~\citep{liu2021IMTL, navon2022NashMTL, lin2023DBMTL}. During training, all input images are resized to $288\times 384$, and we set the batch size to $8$. The experiments are implemented with PyTorch and executed on NVIDIA A100-80G GPUs.

\vspace{-0.55em}
\paragraph{Cityscapes Dataset}
The implementations for Cityscapes benchmark demonstrate substantial alignment with the one on NYUv2~\citep{lin22RLW, lin2023DBMTL}. Specifically, we adopt the identical DeepLabV3+~\citep{chen2018DeepLabV3Plus} architecture, leveraging an ImageNet-pretained dilated ResNet 50 network as the backbone, while the ASPP module serves as task-specific decoders. For model optimization, we establish a $200$-epoch training regime utilizing Adam optimizer, with the initial learning rate of $10^{-4}$ and weight decay of $10^{-5}$. The learning rate undergoes a scheduled reduction to $5\times10^{-5}$ upon reaching the $100$-epoch milestone. We maintain consistency of loss functions with NYUv2: cross-entropy loss and $L_1$ loss are employed for semantic segmentation and depth estimation, respectively. We also adopted logarithmic transformation as in previous studies~\citep{liu2021IMTL, navon2022NashMTL, lin2023DBMTL}. Throughout the training process, all input images are resized to $128\times 256$, and we utilize a batch size of $64$. The experiments are implemented with PyTorch on NVIDIA A100-80G GPUs.

\vspace{-0.55em}
\paragraph{Office-Home Dataset}
Building upon established protocols from prior works~\citep{lin22RLW, lin2023DBMTL}, we implement an ImageNet-pretrained ResNet-18 network architecture as the shared backbone, complemented by a linear layer serving as task-specific decoders. In pre-processing, all input images are resized to $224\times 224$. The batch size and the training epoch are set to $64$ and $100$, respectively. The optimization process employs the Adam optimizer with the learning rate of $10^{-4}$ and the weight decay of $10^{-5}$. We utilize cross-entropy loss for all classification tasks, with classification accuracy serving as the evaluation metric. We also adopted logarithmic transformation as in previous studies~\citep{liu2021IMTL, navon2022NashMTL, lin2023DBMTL}. The ``Avg." reported in the main manuscript represents the mean performance gains across three independent tasks, which is notably excluded from the calculation of overall task-level performance gains. The experiments are implemented with PyTorch and executed on NVIDIA A100-80G GPUs.

\vspace{-0.55em}
\paragraph{Office-31 Dataset}
The configurations on Office-31~\citep{saenko2010Office31} dataset exhibit notable parallels with the ones on Office-Home~\citep{lin22RLW, lin2023DBMTL}. Concretely, we deploy a ResNet-18 network architecture pre-trained on the ImageNet dataset as the shared backbone, complemented by task-specific linear layers for classification outputs. The data processing pipeline standardizes input images to $224\times 224$, while the training protocol extends across $100$ epochs with a fixed batch size of $64$. The Adam optimizer configured with a learning rate of $10^{-4}$ and the weight decay of $10^{-5}$ is used. The cross-entropy loss is used for all the tasks and classification accuracy is used as the evaluation metric. We adopted logarithmic transformation as in previous studies~\citep{liu2021IMTL, navon2022NashMTL, lin2023DBMTL}. The ``Avg." reported in the main manuscript represents the mean performance gains across three independent tasks, which is notably excluded from the calculation of overall task-level performance gains. The experiments are implemented with PyTorch and executed on NVIDIA A100-80G GPUs.

\begin{table}[t]
\centering
\caption{Performance on Office-31 dataset with $3$ diverse image classification tasks. $\uparrow$ indicates the higher the metric values, the better the methods' performance. The best and second-best results of each metric are highlighted in \textbf{bold} and \underline{underline}, respectively.}
\label{tab:mtl-office-31}
\vspace{-0.70em}
\resizebox{1.00\linewidth}{!}{
\begin{tabular}{lccccc}
\toprule
Method         & \multicolumn{1}{c}{\textbf{Amazon}} & \multicolumn{1}{c}{\textbf{DSLR}} & \multicolumn{1}{c}{\textbf{Webcam}} & \multicolumn{1}{c}{\textbf{Avg.}$\uparrow$} & \multicolumn{1}{c}{\bm{$\Delta{\mathrm{p}}_{\text{task}}$}${\uparrow}$}  \\ \midrule
Single-Task Baseline & $\textbf{86.61}$ & $95.63$ & $96.85$ & $93.03$ & $0.00$  \\ \midrule
\grow EW & $83.53$ & $97.27$ & $96.85$ & $92.55_{\pm0.62}$ & $\textcolor{resultblue}{-0.61}_{\pm0.67}$   \\
GLS~\citep{chennupati2019GLS} & $82.84$ & $95.62$ & $96.29$ & $91.59_{\pm0.58}$ & $\textcolor{resultblue}{-1.63}_{\pm0.61}$     \\
RLW~\citep{lin22RLW} & $83.82$ & $96.99$ & $96.85$ & $92.55_{\pm0.89}$ & $\textcolor{resultblue}{-0.59}_{\pm0.95}$  \\
UW~\citep{kendall2018UW}  & $83.82$ & $97.27$ & $96.67$ & $92.58_{\pm0.84}$ & $\textcolor{resultblue}{-0.56}_{\pm0.90}$   \\
DWA~\citep{liu2019DWA} & $83.87$ & $96.99$ & $96.48$ & $92.45_{\pm0.56}$ & $\textcolor{resultblue}{-0.70}_{\pm0.62}$    \\
IMTL-L~\citep{liu2021IMTL} & $84.04$ & $96.99$ & $96.48$ & $92.50_{\pm0.52}$ & $\textcolor{resultblue}{-0.63}_{\pm0.58}$   \\
IGBv2~\citep{dai2023IGBv2}  & $84.52$ & $98.36$  & $98.05$ & $93.64_{\pm0.26}$    & $\textcolor{purple}{+0.56}_{\pm0.25}$    \\
\midrule
MGDA~\citep{desideri12MGDA} & $85.47$ & $95.90$ & $97.03$ & $92.80_{\pm0.14}$ & $\textcolor{resultblue}{-0.27}_{\pm0.15}$  \\
GradNorm~\citep{chen2018GradNorm} & $83.58$ & $97.26$ & $96.85$ & $92.56_{\pm0.87}$ & $\textcolor{resultblue}{-0.59}_{\pm0.94}$  \\
PCGrad~\citep{yu2020PCGrad} & $83.59$ & $96.99$ & $96.85$ & $92.48_{\pm0.53}$ & $\textcolor{resultblue}{-0.68}_{\pm0.57}$   \\
GradDrop~\citep{chen2020GradDrop} & $84.33$ & $96.99$ & $96.30$ & $92.54_{\pm0.42}$ & $\textcolor{resultblue}{-0.59}_{\pm0.46}$   \\
GradVac~\citep{wang2021GradVac} & $83.76$ & $97.27$ & $96.67$ & $92.57_{\pm0.73}$ & $\textcolor{resultblue}{-0.58}_{\pm0.78}$   \\
IMTL-G~\citep{liu2021IMTL}  & $83.41$ & $96.72$ & $96.48$ & $92.20_{\pm0.89}$ & $\textcolor{resultblue}{-0.97}_{\pm0.95}$   \\
CAGrad~\citep{liu2021CAGrad} & $83.65$ & $95.63$ & $96.85$ & $92.04_{\pm0.79}$ & $\textcolor{resultblue}{-1.14}_{\pm0.85}$  \\
MTAdam~\citep{malkiel2021MTAdam}  & $85.52$ & $95.62$ & $96.29$ & $92.48_{\pm0.87}$ & $\textcolor{resultblue}{-0.60}_{\pm0.93}$ \\
Nash-MTL~\citep{navon2022NashMTL} & $85.01$ & $97.54$ & $97.41$ & $93.32_{\pm0.82}$ & $\textcolor{purple}{+0.24}_{\pm0.89}$  \\
MetaBalance~\citep{he2022MetaBalance} & $84.21$ & $95.90$ & $97.40$ & $92.50_{\pm0.28}$ & $\textcolor{resultblue}{-0.63}_{\pm0.30}$  \\
MoCo~\citep{fernando2023MoCo}  & $84.33$ & $97.54$ & $98.33$ & $93.39$ & -   \\
Aligned-MTL~\citep{senushkin2023AlignedMTL} & $83.36$ & $96.45$ & $97.04$ & $92.28_{\pm0.46}$ & $\textcolor{resultblue}{-0.90}_{\pm0.48}$    \\
\midrule
IMTL~\citep{liu2021IMTL}  & $83.70$ & $96.44$ & $96.29$ & $92.14_{\pm0.85}$ & $\textcolor{resultblue}{-1.02}_{\pm0.92}$   \\
DB-MTL~\citep{lin2023DBMTL}  & $85.12$ & $\textbf{98.63}$ & $\underline{98.51}$ & $\underline{94.09}_{\pm0.19}$ & $\textcolor{purple}{\underline{+1.05}}_{\pm0.20}$          \\ 
\midrule
\brow Rep-MTL (EW) & $\underline{85.93}$  & $\underline{98.54}$ & $\textbf{98.67}$ & $\textbf{94.38}_{\pm0.53}$ & $\textcolor{purple}{\textbf{+1.31}}_{\pm0.58}$   \\ \bottomrule
\end{tabular}}
\vspace{-0.35em}
\end{table}

\section{Office-31 Image Classification Results}
\label{app:results}
This appendix section provides a thorough discussion of our experimental results on Office-31~\citep{saenko2010Office31} dataset, presenting detailed observations of performance that were omitted from the main text due to space limitations.

As shown in Table~\ref{tab:mtl-office-31}, Rep-MTL achieves the highest overall performance among all compared MTO methods. It obtains an average accuracy (\textbf{Avg.}$\uparrow$) of $94.38\%$ and a total performance gain ($\Delta{\mathrm{p}_\text{task}}\uparrow$) of $+1.31\%$ over the single-task learning (STL) baseline. This result surpasses the next-best method, DB-MTL, which achieves a gain of $+1.05\%$, and stands in stark contrast to the Equal Weighting (EW) baseline that suffers from negative transfer among tasks ($\Delta$ = $-0.61\%$). This demonstrates Rep-MTL's superior ability to effectively manage multi-domain learning on Office-31. 

In addition, task-specific performance reveals several notable findings. First, on both the \textbf{Webcam} and \textbf{Amazon} domains, Rep-MTL achieves competitive accuracies of $98.67\%$ and $85.93\%$, respectively. Its performance on the challenging \textbf{Amazon} domain is particularly noteworthy, outperforming the strong DB-MTL~\citep{lin2023DBMTL} baseline by a significant margin of $+0.81\%$. This improvement is particularly significant due to the varying lighting conditions and image quality. Second, on \textbf{DSLR} domain, Rep-MTL delivers a competitive accuracy of $98.54\%$, narrowly trailing DB-MTL~\citep{lin2023DBMTL} ($98.63\%$) in a tightly contested result.

These results offer key insights into the strengths and limitations of Rep-MTL. On one hand, Rep-MTL demonstrates capabilities to handle multiple tasks effectively, consistently achieving balanced and top-tier performance gains across different tasks. The substantial gains on the \textbf{Amazon} and \textbf{Webcam} tasks more than compensate for the marginal difference on \textbf{DSLR}, leading to the best overall average. On the other hand, however, this balanced approach comes with a trade-off: while Rep-MTL avoids significant performance degradation in task-specific performance compared to existing methods, it may not consistently achieve significant gains across all sub-tasks simultaneously. This observation is particularly evident in the results of the \textbf{DSLR} task on Office-31~\citep{saenko2010Office31} dataset, where Rep-MTL achieves strong but not leading performance.

Overall, the experimental results suggest that while Rep-MTL has successfully advanced the state-of-the-art in challenging multi-task dense prediction benchmarks, there remains scope for further enhancement. Future research directions could focus on developing mechanisms to maintain the current balanced performance with explicit information sharing while pushing the boundaries of task-specific excellence. This could potentially involve exploring more complex cross-task interactions or adaptive optimization strategies that can better leverage task-specific characteristics.

\begin{figure}[t]
    % \vspace{-0.25em}
    \begin{center}
    \includegraphics[width=1.00\linewidth]{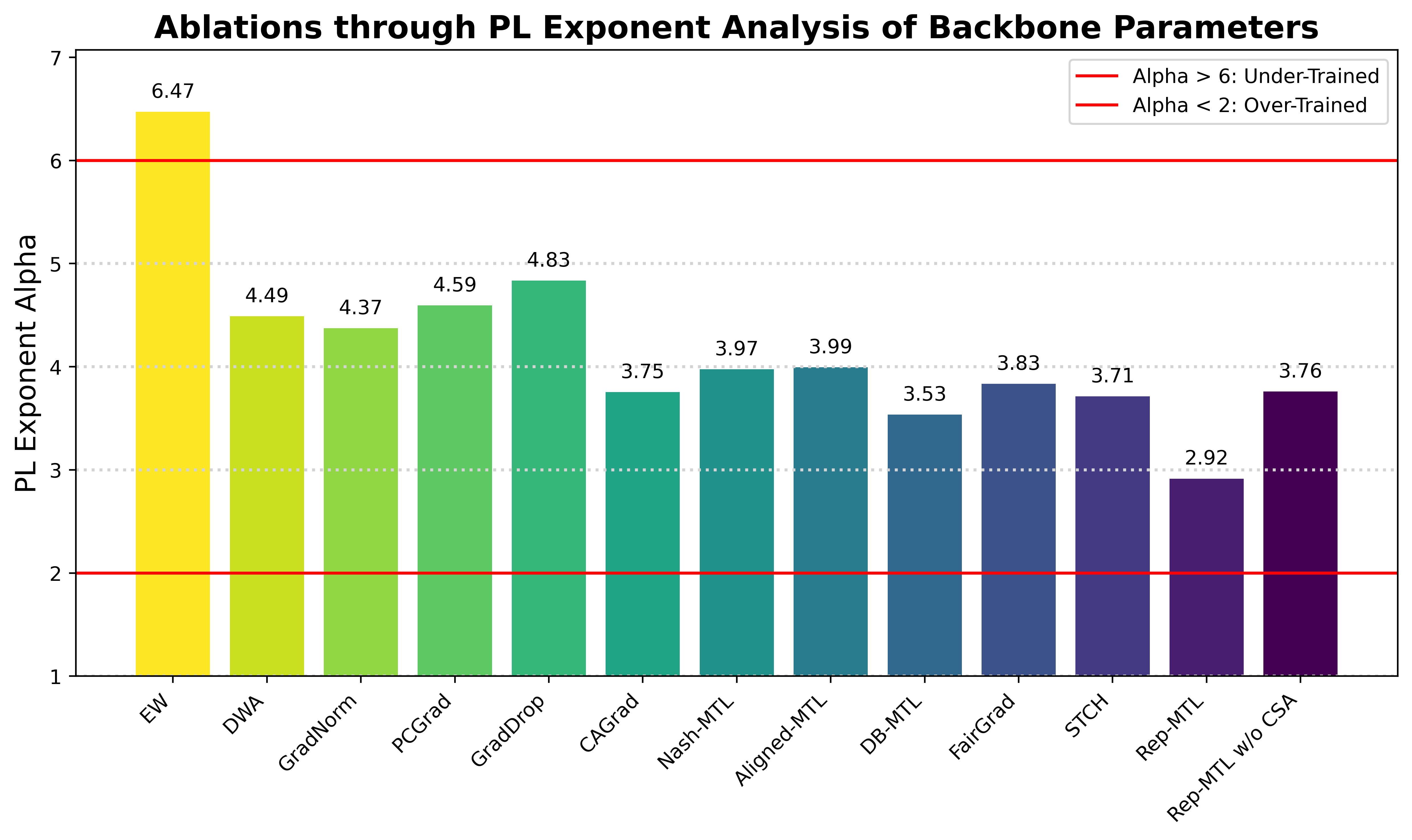}
    \hspace{-0.8em}
    \end{center}
    \vspace{-2.65em}
    \caption{Ablation studies through PL exponent metrics~\citep{ NC2021WeightWatcher} for shared parameters in backbones trained with or without cross-task saliency alignment (notated as ``Rep-MTL w/o CA'') on NYUv2~\citep{silberman2012NYUv2}. The PL exponent alpha quantifies how well the backbone adapts to the overall MTL objectives, where lower values indicate more effective training. Values outside the range $[2,6]$ suggest potential over- or under-training due to the insufficient cross-task positive transfer. We leverage this measurement to validate the effectiveness of the cross-task saliency alignment mechanism in our proposed Rep-MTL, as well-trained backbones suggest beneficial information sharing to the overall MTL objectives.}
    \label{fig:alpha_ablation_backbone}
    \vspace{-0.4em}
\end{figure}

\begin{figure*}[t]
    % \vspace{-0.25em}
    \begin{center}
    \includegraphics[width=1.00\linewidth]{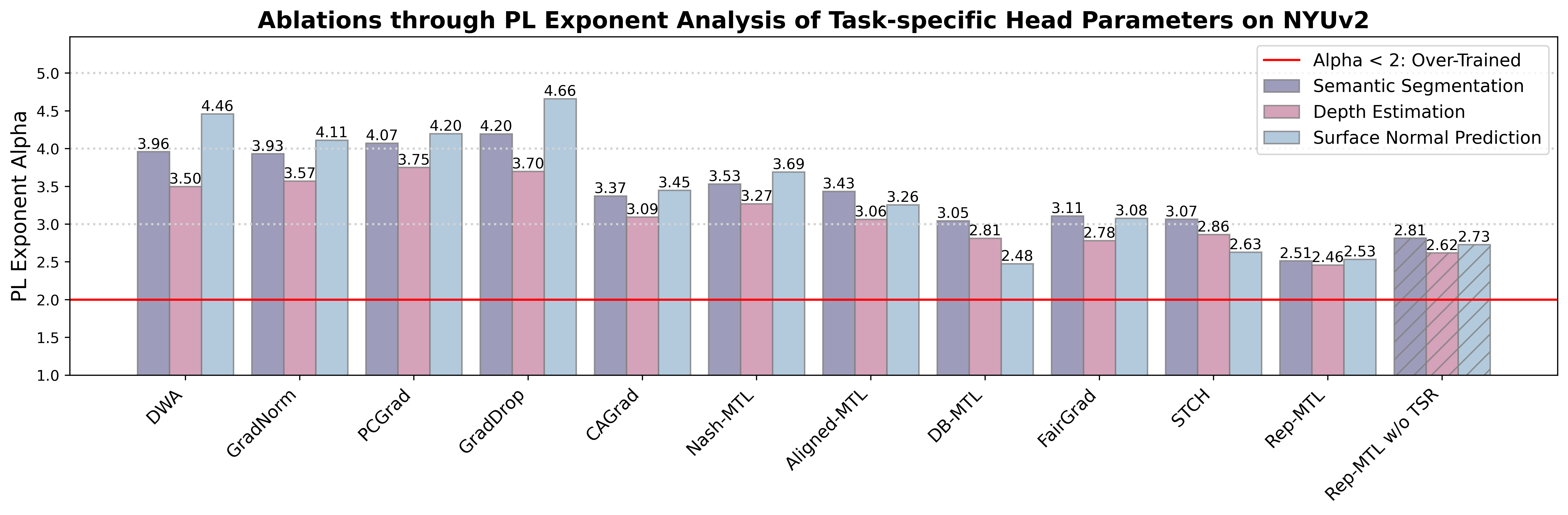}
    \hspace{-1.2em}
    \end{center}
    \vspace{-3.20em}
    \caption{Ablation studies through PL exponent metrics~\citep{JMLR2021selfreg, NC2021WeightWatcher} for parameters in diverse decoders trained with or without task-specific saliency regulation in Rep-MTL (notated as ``Rep-MTL w/o TR'') on NYUv2~\citep{silberman2012NYUv2}. The PL exponent alpha quantifies how well each decoder adapts to its task-specific objective, where lower values indicate more effective training. Values outside the range $[2,6]$ suggest potential over- or under-training due to task conflicts. The variation across different heads of each method indicates training imbalance. We leverage this measurement to validate the effectiveness of task-specific saliency regulation in Rep-MTL, as well-trained decoders should exhibit both \textit{low and balanced} metric values, indicating successful negative transfer mitigation while preserving task-specific information. The results show that task-specific saliency regulation effectively helps task-specific learning and yields superior and more balanced metrics.}
    \label{fig:alpha_ablation_heads}
    \vspace{-0.3em}
\end{figure*}

\section{Ablations with PL Exponent Alpha Metrics}
\label{app:alpha_ablation}
While our analysis in Section~\ref{subsection:exp_ple_analysis} demonstrates Rep-MTL's overall effectiveness in achieving effective multi-task learning—facilitating positive cross-task information sharing while preserving task-specific patterns for negative transfer mitigation—it does not isolate the contributions of individual components. This appendix section thus presents an additional empirical evaluation of Rep-MTL's two key mechanisms: Cross-Task Saliency Alignment (CA) and Task-specific Saliency Regularization (TR). We first introduce the practical implications of this metric, followed by ablation studies examining each component's effectiveness and distinct contribution to Rep-MTL's overall performance.

\subsection{Power Law (PL) Exponent Alpha Analysis}
To rigorously evaluate the effectiveness of Rep-MTL's components beyond commonly-used performance metrics, we employ Power Law (PL) exponent alpha~\citep{JMLR2021selfreg, NC2021WeightWatcher}, a theoretically grounded measure from Heavy-Tailed Self-Regularization (HT-SR) theory~\citep{martin2021implicit, mahoney2019traditionalHTSR}. It provides a systematic framework for analyzing the representation capacity and overall learning quality of deep neural networks. In particular, PL exponent alpha is computed for each layer's weight matrix $W$ by fitting the Empirical Spectral Density (ESD) of its correlation matrix $X = W^TW$ to a truncated Power Law distribution: $\rho(\lambda) \sim \lambda^{-\alpha}$, where $\rho(\lambda)$ denotes the ESD, and $\lambda$ represents eigenvalues of correlation matrix.

Empirical studies have established that well-trained neural networks typically exhibit PL exponent values within the range $\alpha \in [2,4]$. Values outside this range often indicate suboptimal training dynamics: specifically, $\alpha < 2$ suggests insufficient learning, while $\alpha > 6$ indicates potential over-parameterization or training instabilities. This characteristic makes the PL exponent particularly valuable for assessing training effectiveness across different network architectures and optimization strategies.

In the context of multi-task learning, this metric offers unique insights into both cross-task knowledge transfer and task-specific learning patterns. In particular, for shared backbone parameters, lower alpha values (within the optimal range) typically indicate effective cross-task information sharing, suggesting successful optimization toward the overall MTL objectives. For task-specific heads, balanced and moderately low alpha values across different tasks suggest the preservation of task-specific patterns while minimizing negative transfer effects. Built upon this view, we can systematically evaluate how each component in Rep-MTL contributes to achieving optimal MTL dynamics.

\subsection{Effects of Cross-Task Saliency Alignment}
\label{app:ple_ablation_sa}
Similar to the empirical analysis in Section~\ref{subsection:exp_ple_analysis}, we analyze the effectiveness of Cross-Task Saliency Alignment by examining PL exponent alpha of the DeepLabV3+ backbone parameters on NYUv2~\citep{silberman2012NYUv2} dataset.

As shown in Figure~\ref{fig:alpha_ablation_backbone}, models trained with our cross-task alignment mechanism exhibit alpha values within the optimal range of $[2,4]$, indicating well-learned and generalizable model parameters in the shared backbone, comparing models trained with and without this alignment mechanism. This demonstrates the effectiveness of our Cross-Task Saliency Alignment for positive information sharing.

\subsection{Effects of Task-specific Saliency Regulation}
\label{app:ple_ablation_sr}
To evaluate the impact of Task-specific Saliency Regulation, we examine the PL exponent alpha of parameters in the DeepLabV3+ task decoder parameters on NYUv2~\citep{silberman2012NYUv2}.

As illustrated in Figure~\ref{fig:alpha_ablation_heads}, the result reveals that models employing our regulation mechanism demonstrate alpha values consistently within the optimal range and exhibit more balanced values across all task-specific heads. This balanced distribution suggests the successful preservation of task-specific features while avoiding over-specialization or interference between tasks ($2.60, 2.63, 2.45$). In contrast, models trained with Rep-MTL without this regulation mechanism exhibit poor and more dispersed PL exponent alpha across decoders ($2.89, 2.74, 2.59$). This wider variation indicates potential negative transfer and suboptimal task-specific learning. The consistency of alpha values across different task heads in regulated models provides strong evidence that the Task-specific Saliency Regulation in Rep-MTL effectively maintains task-specific patterns.

\begin{figure}[t]
    % \vspace{-0.25em}
    \begin{center}
    \includegraphics[width=1.00\linewidth]{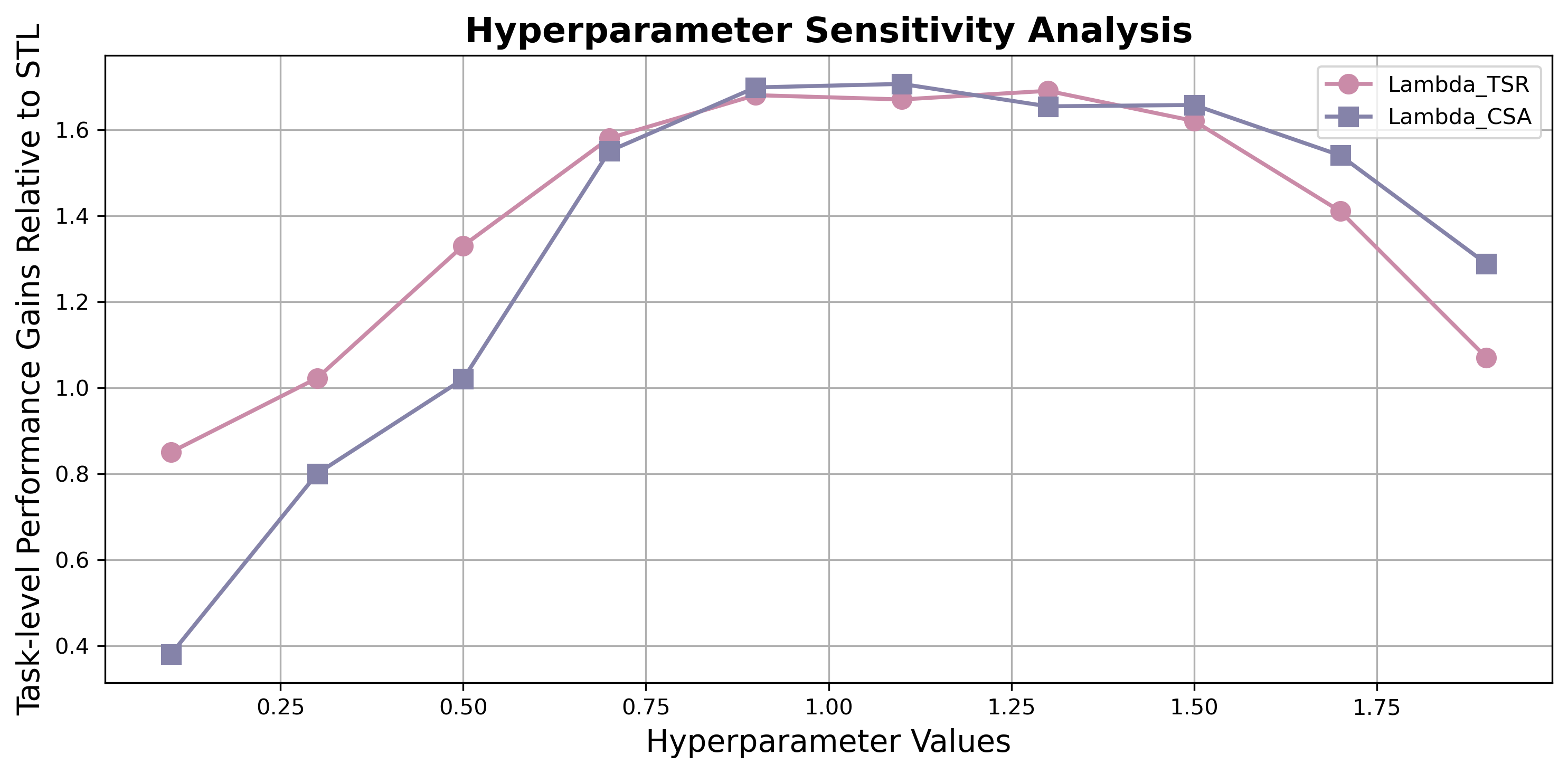}
    \hspace{-0.8em}
    \end{center}
    \vspace{-2.65em}
    \caption{Hyper-parameter sensitivity analysis of our Rep-MTL on NYUv2~\citep{silberman2012NYUv2} dataset. We empirically evaluate the impact of two critical hyper-parameters, $\lambda_{tsr}$ and $\lambda_{csa}$, by fixing one as $\lambda = 0.9$ while varying the other one across a comprehensive range as $\{0.1, 0.3, 0.5, 0.7, 0.9, 1.1, 1.3, 1.5, 1.7, 1.9\}$. The results demonstrate that Rep-MTL maintains stable and competitive performance $\Delta{\mathrm{p}_\text{task}}$ over a substantial range (${0.7, 0.9, 1.1, 1.3, 1.5}$), indicating its robust insensitivity to hyper-parameter variations.}
    \label{fig:hyperparam}
    \vspace{-0.4em}
\end{figure}

\section{Additional Empirical Analysis}
\label{app:empirical}
This appendix section presents an empirical investigation designed to further validate the effectiveness and robustness of Rep-MTL. We conduct empirical analyses of hyper-parameter sensitivity and computational efficiency to provide insights into the practical deployment considerations.

\subsection{Analysis of Hyper-parameter Sensitivity}
\label{app:hyper}
We systematically evaluate Rep-MTL's sensitivity to its two primary hyper-parameters, $\lambda_{tsr}$ and $\lambda_{csa}$, on the NYUv2~\citep{silberman2012NYUv2} dataset. 
Figure~\ref{fig:hyperparam} illustrates the task-level performance gains relative to STL baselines ($\Delta{\mathrm{p}_\text{task}}$) across various hyper-parameter configurations. Our analysis involves fixing one hyper-parameter at $0.9$ while varying the other one across a comprehensive range: $\{0.1, 0.3, 0.5, 0.7, 0.9, 1.1, 1.3, 1.5, 1.7, 1.9\}$. For example, when evaluating the sensitivity of hyper-parameter $\lambda_{tsr}$, when fixing the $\lambda_{csa} = 0.9$ then conduct a series of experiments. All experiments are conducted on NVIDIA A100-80G GPUs to ensure consistent evaluation conditions. 

The results reveal several key findings: First, Rep-MTL demonstrates great stability across a wide range of hyper-parameter combinations, particularly within the range of $\{0.7, 0.9, 1.1, 1.3, 1.5\}$ for both $\lambda_{tsr}$ and $\lambda_{csa}$. Second, the method consistently achieves positive performance gains ($\Delta{\mathrm{p}_\text{task}} > 0$) across most hyper-parameter settings, indicating robust improvement over STL baselines. Third, Cross-task Saliency Alignment (CSA) in Rep-MTL, controlled by $\lambda_{csa}$, acts as a crucial component. While small values of $\lambda_{csa}$ lead to suboptimal performance, increasing it beyond a certain threshold demonstrates a significant impact on the overall MTL performance. Based on these observations, we conducted grid search over $\{0.7, 0.9, 1.1, 1.3, 1.5\}$ for both
$\lambda_{tsr}$ and $\lambda_{csa}$ to determine optimal configurations for all datasets in this paper.

\begin{figure}[t]
    % \vspace{-0.25em}
    \begin{center}
    \includegraphics[width=1.00\linewidth]{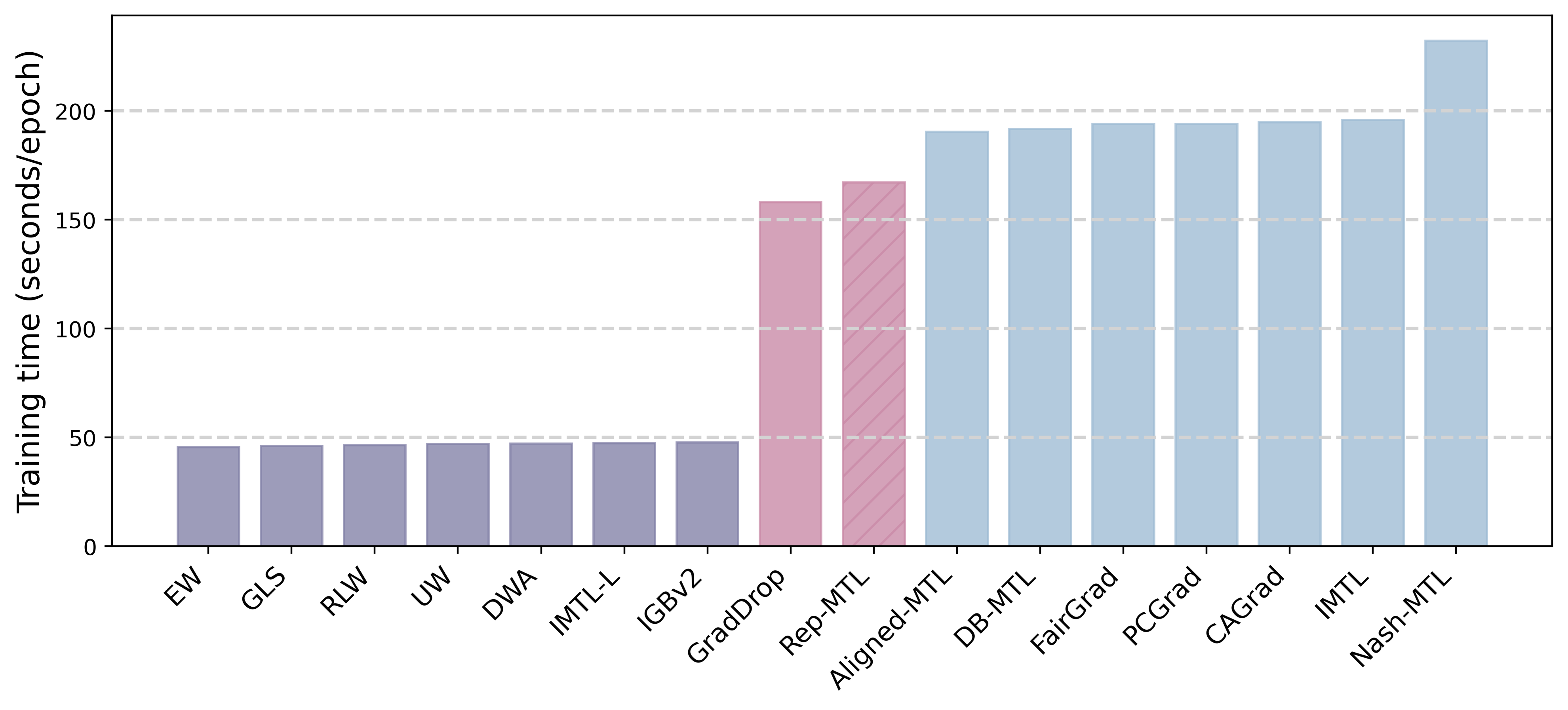}
    \hspace{-0.8em}
    \end{center}
    \vspace{-2.65em}
    \caption{Training time per epoch comparison across different MTL optimization methods on NYUv2~\citep{silberman2012NYUv2}. Methods are categorized into three training efficiency tiers (indicated by different colors), highlighting the inherent trade-off between computational speed and optimization effectiveness in MTL scenarios.}
    \label{fig:runtime}
    \vspace{-0.4em}
\end{figure}

\subsection{Analysis of Training Time}
\label{app:runtime}
To further evaluate the efficiency of Rep-MTL, we conduct a runtime empirical analysis on NYUv2~\citep{silberman2012NYUv2} dataset. Figure~\ref{fig:runtime} presents the average per-epoch training time across different MTL optimization methods, with all experiments conducted over 100 epochs on NVIDIA A100-80G GPUs. Our analysis reveals that Rep-MTL achieves a comparatively favorable balance between training speed and optimization effectiveness. While it requires more training resources than loss scaling methods due to the computation of task saliencies as task-specific gradients in the representation space, it demonstrates superior efficiency compared to most gradient manipulation methods. This increased cost is inherent to approaches requiring second-order (gradient) information, representing a fundamental trade-off and room for further improvement in MTL optimization.

\subsection{Analysis of Learning Rate Scaling}
\label{app:learning_rate}
\begin{figure}[t]
    % \vspace{-0.25em}
    \begin{center}
    \includegraphics[width=1.00\linewidth]{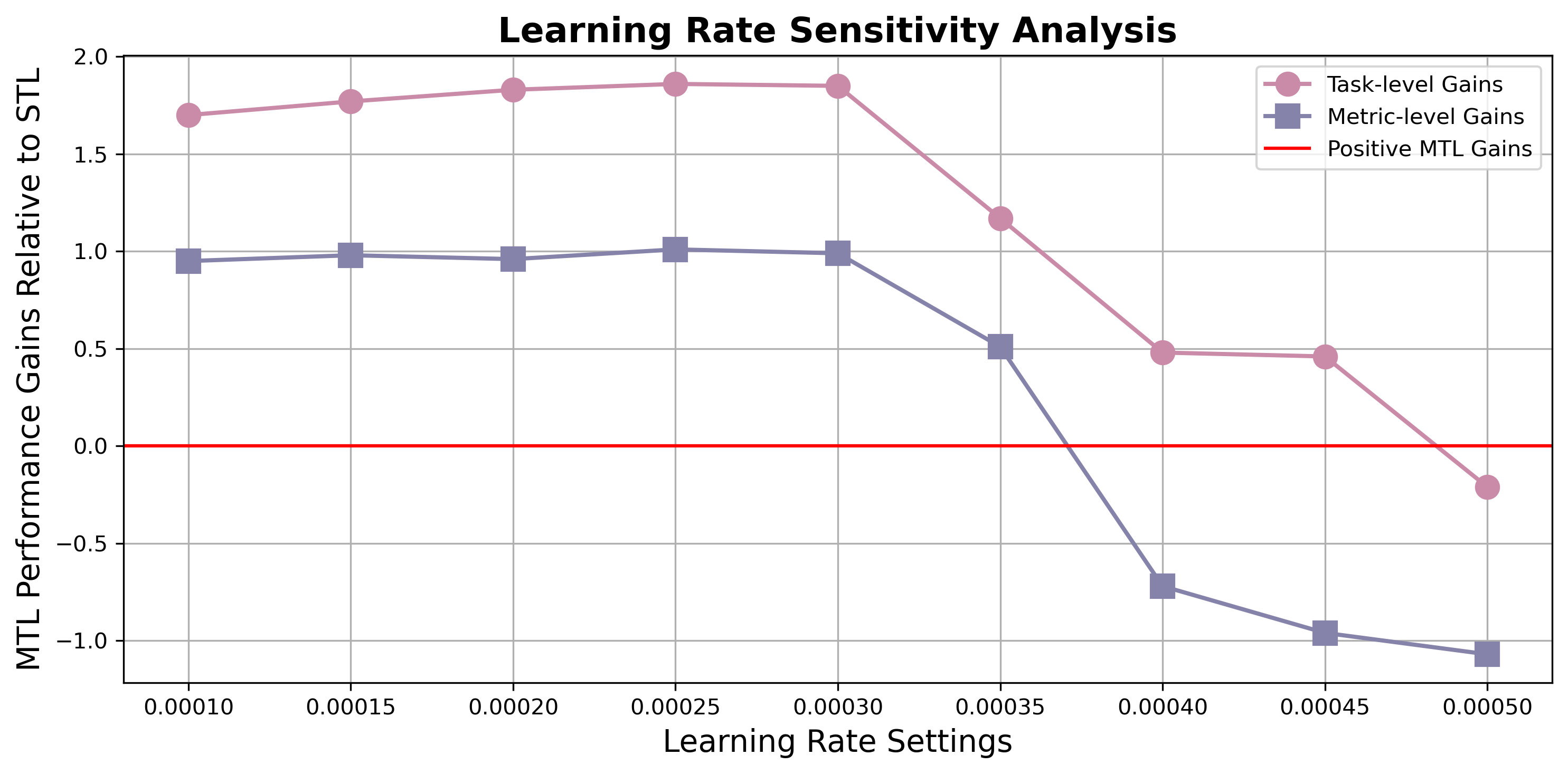}
    \hspace{-0.8em}
    \end{center}
    \vspace{-2.65em}
    \caption{Learning rate sensitivity analysis of our proposed Rep-MTL on NYUv2~\citep{silberman2012NYUv2} dataset. To evaluate the impact of learning rate variations, we systematically scale the learning rate from the default benchmark setting of $1e-4$ to $5e-4$, using a step size of $5e-5$. For each setting, we report the task-level ($\Delta{\mathrm{p}_\text{task}}$) and metric-level ($\Delta{\mathrm{p}_\text{metric}}$) performance gains. Each experiment is repeated three times. The results show that Rep-MTL maintains stable and competitive $\Delta{\mathrm{p}_\text{task}}$ and $\Delta{\mathrm{p}_\text{metric}}$ over a substantial range, indicating its favorable robustness to learning rate variations.}
    \label{fig:learning_rate_sensitivity}
    \vspace{-0.4em}
\end{figure}

Recent studies~\citep{xin2022current} suggest that different choice of learning rate may impose a strong impact on MTO methods performance. To further demonstate Rep-MTL's robustness, we conduct experiment of learning rate sensitivity on NYUv2~\citep{silberman2012NYUv2} with diverse learning rate settings, as illustrated in Figure~\ref{fig:learning_rate_sensitivity}. Specifically, we scale the learning rate from the default benchmark setting of $1e-4$ to $5e-4$ with a step size of $5e-5$. For each setting, we measure the task-level ($\Delta{\mathrm{p}_\text{task}}$) and metric-level ($\Delta{\mathrm{p}_\text{metric}}$) performance gains. The results show that Rep-MTL maintains stable and competitive $\Delta{\mathrm{p}_\text{task}}$ and $\Delta{\mathrm{p}_\text{metric}}$ over a substantial range, indicating its favorable robustness to learning rate variations.

\end{document}